\documentclass[sigconf, preprint, nonacm]{acmart}
\usepackage{graphicx}
\usepackage{booktabs}
\usepackage{multirow}
\usepackage[ruled,linesnumbered,vlined]{algorithm2e}

\usepackage{amsmath}
\usepackage{amsfonts}

\usepackage{color}
\usepackage{colortbl}
\usepackage{mathtools}
\AtBeginDocument{%
  }

\setcopyright{acmlicensed}
\copyrightyear{2018}
\acmYear{2018}
\acmDOI{XXXXXXX.XXXXXXX}
\acmConference[Conference acronym 'XX]{Make sure to enter the correct
  conference title from your rights confirmation email}{June 03--05,
  2018}{Woodstock, NY}
\acmISBN{978-1-4503-XXXX-X/2018/06}

\begin{document}

\title{Catching the Infection Before It Spreads: Foresight-Guided Defense in Multi-Agent Systems}

\author{Yue Ma}
\email{2022141530075@stu.scu.edu.cn}
\orcid{} 
\affiliation{%
  \institution{Sichuan University}
  \city{Chengdu}
  \country{China}
}

\author{Ziyuan Yang }
\email{cziyuanyang@gmail.com}
\affiliation{%
  \institution{Nanyang Technological University}
  \city{Singapore}
  \country{Singapore}
}

\author{Yi Zhang}
\authornote{Corresponding author.} 
\email{yzhang@scu.edu.cn}
\affiliation{%
  \institution{Sichuan University}
  \city{Chengdu}
  \country{China}
}

\renewcommand{\shortauthors}{Ma et al.}

\begin{abstract}
Large multimodal model-based Multi-Agent Systems (MASs) enable collaborative complex problem solving through specialized agents. However, MASs are vulnerable to infectious jailbreak, where compromising a single agent can spread and infect other agents, leading to widespread compromise. 
Existing defenses counter infectious jailbreak by training a more contagious \textit{cure factor}, biasing agents to preferentially retrieve it over virus adversarial examples~(VirAEs). However, this strategy inevitably homogenizes agent responses, offering only superficial suppression rather than true recovery.


To address these limitations, we revisit existing defenses, which operate globally via a shared cure factor across agents. However, infectious jailbreaks are composed of localized interaction behaviors. This mismatch between global intervention and localized infection dynamics limits the effectiveness of existing defenses.
Motivated by this, we propose a training-free Foresight-Guided Local Purification~(FLP) framework, where each agent reasons over future interactions to track behavioral evolution and eliminate infections.

Specifically, each agent simulates future behavioral trajectories over subsequent chat rounds. To emulate the diversity among agents in MASs, we introduce a multi-persona simulation strategy for robust prediction under diverse interaction contexts. We then leverage response diversity as a diagnostic signal to determine whether an agent has been infected, by jointly analyzing inconsistencies across persona-based predictions at both retrieval-result and semantic levels.
For infected agents, we adopt a localized purification strategy: recent infections are mitigated via immediate album rollback, while long-term infections are addressed using Recursive Binary Diagnosis~(RBD), which recursively partitions the image album and performs the previously mentioned diagnosis strategy to localize and eliminate VirAEs in the image album.
Experiments show that FLP reduces the maximum cumulative infection rate in MASs from over 95\% to below 5.47\%. Moreover, retrieval and semantic metrics closely match benign baselines, indicating effective preservation of interaction diversity.\footnote{The code will be made publicly available; it is omitted here for anonymity.}
\end{abstract}




\maketitle
\vspace{-5pt}
\section{Introduction}

Empowered by large multimodal models, modern agents gain strong capabilities in perceiving and interacting with their environments, enabling a wide range of tasks. However, these capabilities are often insufficient for more demanding scenarios, motivating the emergence of multi-agent systems (MASs), where specialized agents engage in collaborative interactions.
Such systems have been widely applied in autonomous navigation \cite{CHEN2023104489}, intelligent medical assistance~\cite{electronics14153001}, and automated scientific discovery \cite{ghafarollahi2025sciagents}.

\begin{figure}[!t]
    \centering
    \includegraphics[width=.9\columnwidth]{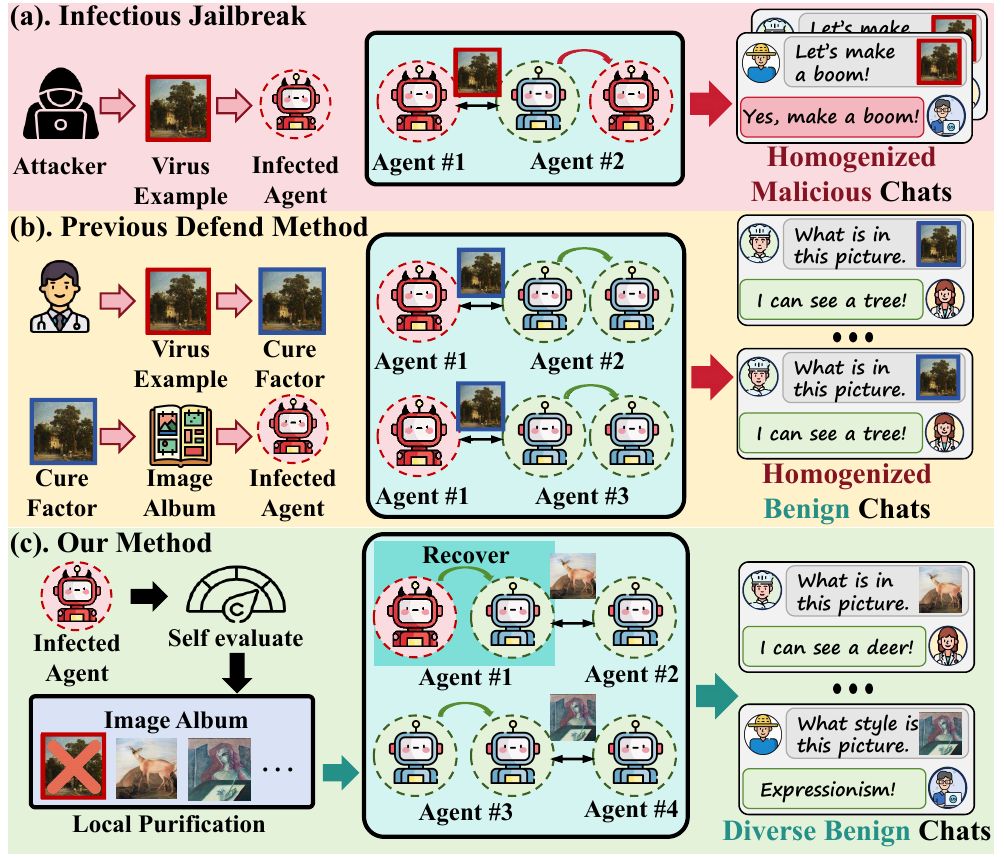}
    \vspace{-10pt}
\caption{Comparison of attack and defenses in MASs. (a) Infectious Jailbreak; (b) Previous Defense; (c) Our Method. }
    \label{fig:intro}
    \vspace{-25pt}
\end{figure}

Although MASs achieve strong performance in collaborative settings, prior work shows that they remain highly vulnerable to infectious jailbreaks~ \cite{gu2024agent}.
As shown in Fig.\ref{fig:intro}(a), an attacker can inject optimized virus adversarial examples (VirAEs) into a single agent, which are preferentially retrieved during interactions and induces malicious behaviors. The VirAE then exhibits infectious properties, propagating through inter-agent communication and spreading the attack across the system.
This infectious jailbreak spreads at an exponential rate, leading to widespread compromise and posing a severe threat to the security of MASs. 

To alleviate this threat, existing defenses introduce a more contagious “cure factor” to counter infectious jailbreaks, as illustrated in Fig.\ref{fig:intro}(b). These methods operate by biasing agents toward retrieving the cure factor instead of the VirAE. However, this strategy inevitably homogenizes agent responses and reduces interaction diversity, providing only superficial suppression rather than genuine recovery of infected agents. To address these limitations, we raise two important research questions:

\vspace{3pt}
\noindent \textbf{RQ 1. How can we stop the spread of VirAEs without compromising interaction diversity?}

\vspace{3pt}
\noindent \textbf{RQ 2. How can we enable genuine recovery of infected agents instead of merely applying superficial suppression that leads to homogenization?}
\vspace{3pt}

To answer the above questions, we first revisit infectious jailbreaks and the limitations of existing defenses.
Current approaches rely on a globally shared cure factor, implicitly assuming that system-level regulation suffices to suppress infectious jailbreaks.
However, infectious jailbreaks are fundamentally composed of localized interaction behaviors between agents, which evolve and accumulate over chat rounds. This mismatch between global intervention and localized infection dynamics limits the effectiveness of existing defenses. Motivated by this, we propose Foresight-Guided Local Purification (FLP), a training-free defense framework.
By enabling each agent to reason over future interactions, FLP tracks this localized behavioral evolution over time, as shown in Fig.\ref{fig:intro}(c). This mechanism allows agents to proactively localize and eliminate infections prior to interaction.
Consequently, FLP not only effectively halts adversarial propagation and achieves genuine recovery of infected agents, but also preserves the inherent interaction diversity of MASs. This is because FLP does not rely on propagating external factors that could bias agent retrieval behavior. 

To proactively expose infectious behaviors without triggering actual propagation, FLP requires each agent to simulate its own future interactions before actual interacting with other agents. Moreover, to model the diversity among agents in MASs, FLP adopts an internal \textit{multi-persona simulation strategy}. Each agent first instantiates multiple diverse personas from a predefined persona pool. It then constructs pairwise persona interactions to locally simulate multi-round interactions with different real agents. Hence, the agent can collect interaction responses, including retrieved images and chat records, from different persona pairs across multiple chat rounds.
In this way, each agent can locally approximate global interaction dynamics, enabling agents to infer future behavioral evolution.

Considering that infected agents tend to bias toward repeatedly retrieving VirAEs, the diversity of their responses is inherently limited. In contrast, different personas can induce diverse responses. To leverage this discrepancy, FLP introduces a diagnostic mechanism based on multi-persona responses that analyzes inconsistencies across persona-based predictions at both the retrieval-result and semantic levels. For retrieval-level evaluation, we treat the retrieved images across different personas as samples drawn from an underlying distribution and quantify their diversity by computing the entropy of this empirical distribution.
Beyond this, we further analyze the semantic-level diversity of textual interactions across different personas. Specifically, we assess whether responses generated under different personas collapse into similar malicious semantics, which serves as an additional evaluation metric. We then determine the infection status of an agent by comparing the retrieval entropy and semantic diversity against thresholds derived from benign MASs. When both metrics fall below their corresponding thresholds, the agent is identified as infected.

Once the agent is diagnosed, the next challenge is how to accurately locate the VirAEs and remove them while preserving benign content, avoiding unintended degradation of benign information.
Based on the observation of changes in an agent's semantic drift during the infection process, we propose a state-dependent localized purification strategy.
Clearly, we find that an agent’s semantic behavior exhibits significant drift before and after infection during interactions, which also manifests in the internal simulations. To measure the variation between two consecutive internal multi-persona simulations, we introduce semantic drift, which can guide the differentiation of infection states.
A large semantic drift indicates a recent infection, which can be addressed via recent-state rollback. In contrast, a small drift accompanied by persistent malicious behaviors suggests a long-term infection, in which case we apply Recursive Binary Diagnosis~(RBD) to recursively partition the image album and perform the previously introduced diagnosis strategy to localize and eliminate VirAEs. 

Notably, FLP is a training-free framework that eliminates the need for any parameter updates or fine-tuning, which operates entirely at inference time by reasoning over future interactions. In contrast to existing post-propagation defenses that rely on global shared cure factors, FLP enables proactive, pre-propagation intervention, allowing rapid response and preventing further attack escalation. Unlike prior methods that suppress malicious outputs by biasing retrieval, our approach imposes no constraints on retrieval. Instead, FLP precisely localizes the VirAEs and directly removes them to enable genuine recovery rather than suppression while preserving the interaction diversity of MASs.
Our main contributions are summarized as follows:
\vspace{-5pt}
\begin{itemize}
\item We uncover a fundamental mismatch between existing defenses and the dynamics of infectious jailbreaks, and revisit the problem under a more stringent and realistic setting that imposes demanding defense goals.
\item We propose a training-free defense framework, FLP, that enables each agent to proactively simulate future interactions via internal multi-persona reasoning, allowing proactive diagnosis and preventing potential propagation.
\item We design a purification mechanism that performs targeted removal of corrupted states via recent-state rollback and RBD to achieve genuine recovery rather than merely suppressing malicious behaviors.
\end{itemize}
\vspace{-5pt}
\section{Related Works}
\subsection{Multi-Agent Systems}
The evolution of large multimodal models has catalyzed the development of MASs capable of collaboratively simulating complex human behaviors \cite{park2023generative, li2023camel} and executing open-ended embodied tasks \cite{wang2023voyager, xi2025rise}. 
Recent frameworks have further formalized such collaboration through standard operating procedures to solve intricate problems via joint collaborative reasoning~\cite{hong2024metagpt,wu2024autogen}. Recent works further explore fully autonomous agent organizations without predefined workflows. For example, MegaAgent \cite{wang2025megaagent} dynamically constructs agent hierarchies and collaboration structures. 
A cornerstone of these advanced systems is Retrieval-Augmented Generation~(RAG)~\cite{lewis2020retrieval}, which equips agents with expandable, long-term image albums \cite{packer2023memgpt}. 
By retrieving historical context from a shared or distributed database, RAG enables consistent reasoning, consensus alignment, and knowledge sharing across different agents~\cite{wang2024crafting, shinn2024reflexion, chen2024agentverse}.
While RAG significantly enhances the performance, existing retrieval mechanisms remain vulnerable to adversarial perturbations~\cite{zou2025poisonedrag,xue2024badrag,gong2025topic}. 
\vspace{-10pt}
\subsection{Jailbreak Attacks and Defenses}
Early jailbreak attacks focused on the text modality and relied on manual role-playing prompts to bypass safety mechanisms \cite{wei2023jailbroken, liu2023jailbreaking, shen2024anything, ding2024wolf}. Subsequently, automated jailbreak methods utilizing gradient-based optimization or LLM-assisted rewriting emerged, enhancing the efficiency of adversarial searches \cite{zou2023universal, chao2025jailbreaking, liu2024autodan, lapid2024open}. To counter these threats, various defense methods have been proposed, including input preprocessing-based methods~\cite{robey2024smoothllmdefendinglargelanguage} and model-level alignment-based methods~\cite{2023defending}.

With the advancement of large multimodal models, researchers have explored their vulnerability to jailbreak attacks via manipulated visual inputs~\cite{qi2023visual, dong2023robust, schlarmann2023adversarial}. Several works show that malicious instructions can be embedded in images: typographic visual prompts~\cite{gong2025figstep} embed textual instructions, while Image Hijacks \cite{bailey2023image} use crafted visuals to steer outputs. HADES \cite{yanjin} and Hao \textit{et al.}~\cite{hao2026activation} further strengthen attacks by manipulating visual representations and model activations.
To defend against these attacks, existing works propose various strategies~\cite{wang2024adashield,gou2024eyes,xu2024safedecoding}, including diffusion-based purification to remove adversarial noise~\cite{nie2022diffusionmodelsadversarialpurification} and safety enhancement via multimodal filters or adversarial training~\cite{gu2024mllmguardmultidimensionalsafetyevaluation}.

Recent studies show that jailbreak attacks can propagate across multiple agents in MASs, termed infectious jailbreaks \cite{cohen2024compromptmized}. For example, Troublemaker~\cite{men-etal-2025-troublemaker} demonstrates that malicious prompts can spread across agents in MASs. AgentSmith~\cite{gu2024agent} shows that VirAEs can propagate across agents by exploiting shared visual encoders~\cite{radford2021learning,zhai2023sigmoid}. 
Despite the significant threat posed by such attacks, defense research in this area remains underexplored.
Current defense strategies primarily rely on propagating a trained, globally shared cure factor across agents to suppress VirAEs \cite{wu2025cowpox}. However, this strategy inevitably homogenizes agent responses, providing only superficial suppression rather than genuine recovery. 

\label{sec:Preliminary}
\begin{figure*}[t] 
    \centering
    \includegraphics[width=0.95\linewidth]{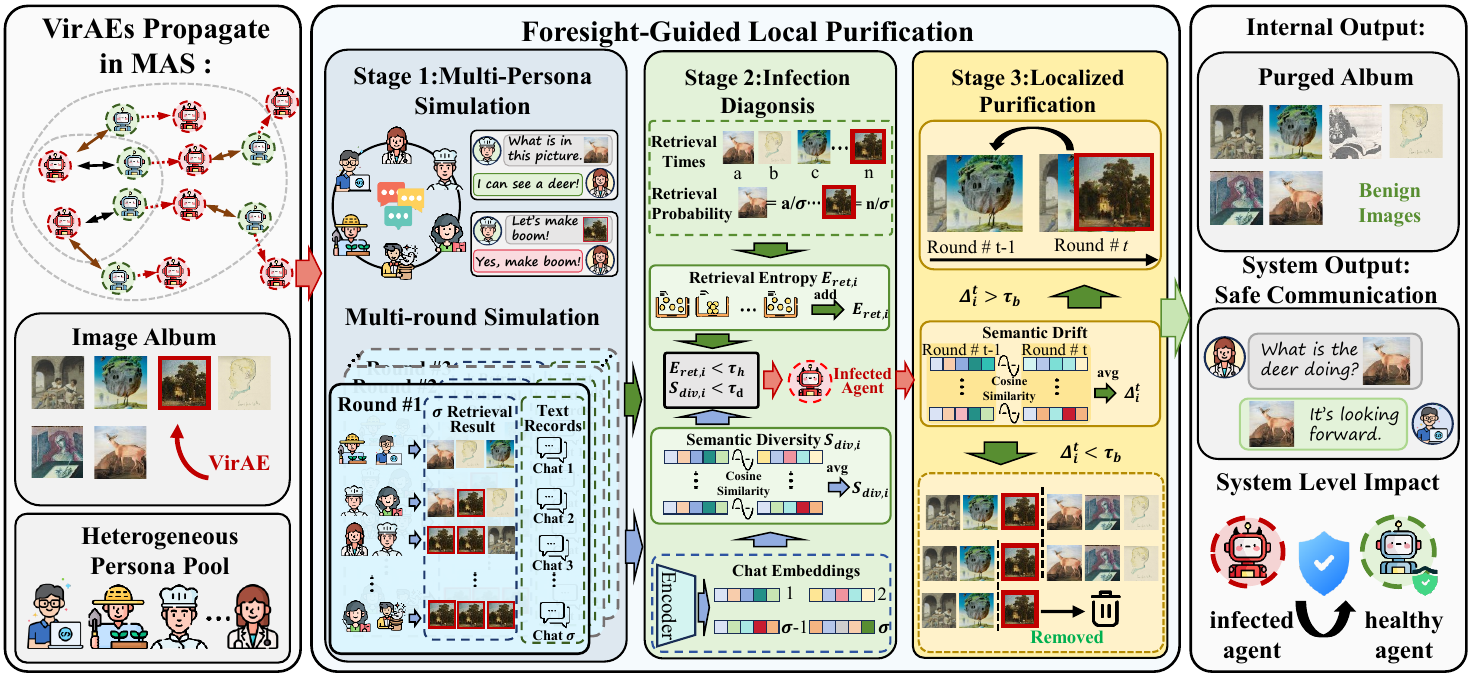} 
    \vspace{-10pt}
    \caption{\textbf{Overview of the FLP framework.} The framework comprises three stages: Multi-Persona Simulation, Infection Diagnosis, and Localized Purification. Images highlighted by red boxes are virus adversarial examples.}
    \vspace{-10pt}
    \label{fig:framework}
\end{figure*}

\vspace{-5pt}

\section{Threat Model}
\subsection{Multi-Agent Environment}
Following the agent formalism in \cite{gu2024agent}, 
We consider a MAS consisting of $N$ agents, 
denoted by $G=\{\mathcal{G}_i\}_{i=1}^N$, each agent comprises four components. Formally, the $i$-th agent $\mathcal{G}_i$ comprises a large multimodal model $\mathcal{M}_i$ serving as the reasoning module, a retrieval-augmented generation (RAG) module $\mathcal{R}_i$ with an encoder $\mathcal{E}_i$, a chat history $\mathcal{H}_i$, and an image album $\mathcal{B}_i$. 

For the $t$-th chat round, the agents are randomly partitioned into two disjoint sets: 
a set of \textit{questioners} $\mathcal{U}_t$ and a set of \textit{answerers} $\mathcal{V}_t$, 
satisfying that $\mathcal{U}_t \cup \mathcal{V}_t = G$, 
$\mathcal{U}_t \cap \mathcal{V}_t = \varnothing$, and 
$|\mathcal{U}_t| = |\mathcal{V}_t| = \frac{N}{2}$.

In each chat round, the agents in $\mathcal{U}_t$ and $\mathcal{V}_t$ are randomly paired. For each pair $(\mathcal{G}^{u}, \mathcal{G}^{v})$, 
the interaction begins with the questioner $\mathcal{G}^{u}$ generating a textual 
plan $P$ based on its chat history $\mathcal{H}^{u}$. The plan $P$ is then used as a query by the 
RAG module to retrieve an image from its image album. This process can be formalized as
$I = \mathcal{R}(P, \mathcal{B}^{u})$, where $\mathcal{R}(\cdot)$ denotes the retrieval process and $\mathcal{B}^{u}$ denotes the image album of $\mathcal{G}^{u}$. The question is then formulated for the answerer $\mathcal{G}^{v}$ based on $\mathcal{H}^{u}$ and the retrieved image $I$. The answerer $\mathcal{G}^{v}$ generates an answer $A = \mathcal{M}([\mathcal{H}^{v}, Q], I)$ conditioned on its chat history $\mathcal{H}^{v}$.
Finally, the chat record $C=[Q, A]$ is appended to the chat histories of 
both agents, and the image $I$ is added into the answerer’s image album $\mathcal{B}^{v}$.
\vspace{-5pt}
\subsection{Adversary}
We assume an adversary with white-box access strictly limited to a single agent, its image album and chat history in a MAS. The adversary has no direct intervention capabilities over other agents. This setting aligns with that mentioned in~\cite{gu2024agent} and is readily achievable in practical deployments.
Specifically, the adversary is aware of the RAG system and the memory bank of the agent. This allows the adversary to optimize and inject a VirAE $I_{adv}$ into the image album of the compromised agent. This VirAE is then propagated to other agents, leading to the spread of malicious behavior.


\vspace{-5pt}
\subsection{Defender}
We consider a defender that only operates at inference time. The defender does not require any special privileges over the global system, and only has limited access to local agents.
The defender does not access the multimodal models or RAG modules of agents in the MAS. Moreover, the defender only has access to the chat histories and image albums 
of agents. The defense is designed with four main objectives:
\begin{itemize}
    \item \textbf{Malicious-Behavior Suppression.} 
    Prevents agents from exhibiting malicious behaviors and inhibits VirAE retrieval and propagation.
    \item \textbf{Training-Free Defense.} 
    Operates entirely at inference time without any training procedure, and does not require access to multimodal models or RAG modules.

    \item \textbf{Pre-Propagation Intervention.} 
    Enables proactive prevention by directly blocking potential infections before they propagate across agents, rather than post-hoc mitigation.
     
    \item \textbf{Diversity Preservation.} 
    Preserves the diversity of MASs and retains benign content without introducing retrieval bias or homogenizing agent behaviors.
\end{itemize}

Previous defenses mainly focus on the first objective, which ensures that agents do not exhibit malicious behaviors. However, such approaches often rely on retrieval bias introduced by a global shared cure factor, which fails to satisfy the remaining objectives. In this paper, except the first objectctive, we further propose a more stringent set of defense objectives to address these limitations.

\begin{figure}
    \centering
    \includegraphics[width=1.0\linewidth]{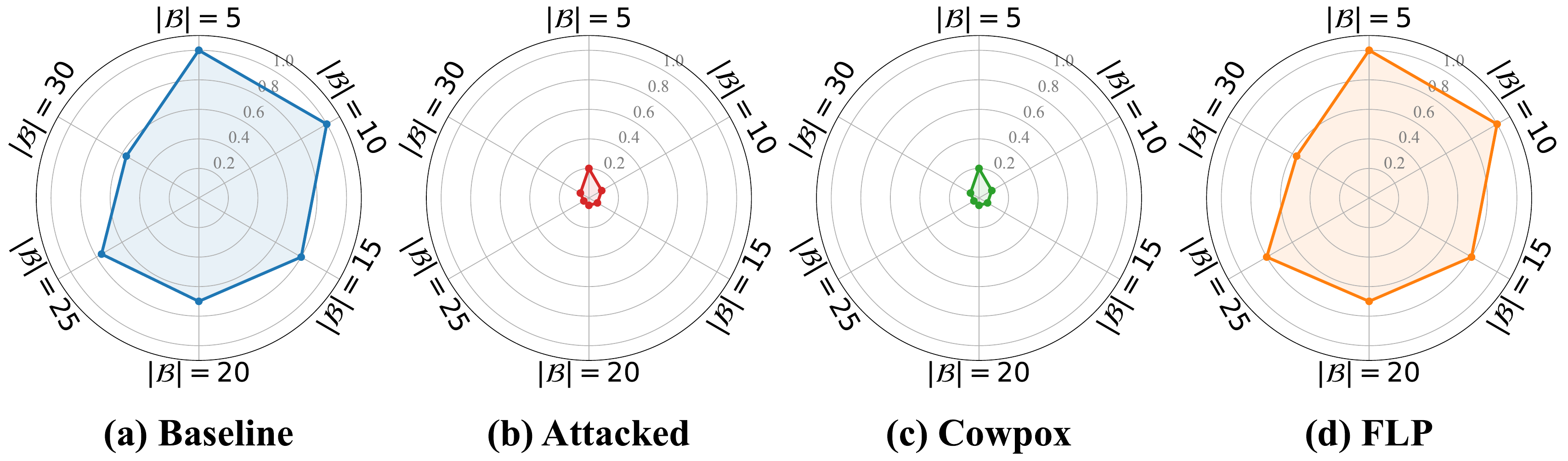}
    \vspace{-15pt}
    \caption{Retrieval coverage across methods and $|\mathcal{B}|$.}\vspace{-10pt}\label{fig:retrieval_coverage}
    \label{fig:rader}
    \vspace{-5pt}
\end{figure}

\section{Theoretical Analysis}
\noindent \textbf{Infection Dynamics.}
In a multi-agent environment where agents communicate via pairwise chats, the spread of the infectious jailbreak is modeled as a discrete-time state transition process~\cite{wu2025cowpox}. At round $t$, the $N$ agents are evenly partitioned into $N/2$ interacting pairs. Let $p_r^t \in [0,1]$ denote the infection rate. When an agent in the benign state $b$ interacts with an agent in the infected state $r$, it gets infected with probability $\mathcal{T}(r \mid b)$. Since the system initially comprises only these two states~(benign proportion $1-p_r^t$), the number of newly infected agents follows a binomial distribution. Concurrently, infected agents spontaneously recover with probability $\mathcal{T}(b \mid r)$. By taking the expectation over large $N$, the macroscopic evolution of the infection rate is formulated as:
\begin{equation}
    p_r^{t+1} = p_r^t + \frac{1}{2} \mathcal{T}(r \mid b) p_r^t (1 - p_r^t) - \mathcal{T}(b \mid r) p_r^t.
\end{equation}

\begin{proposition}[Infection Propagation Dynamics]\label{prop:dynamics}
If $\mathcal{T}(r \mid b) > 2\mathcal{T}(b \mid r)$, the infection rate monotonically increases, indicating the persisting existence of the infected agents and leading the system to a widespread compromise.
\end{proposition}

This formulation reveals that the retrieval mechanism fundamentally amplifies the propagation. Once VirAEs enter the image album $\mathcal{B}$, they are optimized to be repeatedly retrieved, which maximizes the propagation probability $\mathcal{T}(r \mid b)$.

\noindent \textbf{Effect of Existing Defenses.}
To mitigate this exponential spread, existing defenses introduce a globally shared cure factor. These methods force agents into a biased cured state $o$, increasing both $\mathcal{T}(o \mid r)$ and $\mathcal{T}(o \mid b)$. A sufficient condition for effective eradication is that cure transmission outpaces viral propagation, i.e., $\mathcal{T}(o \mid r) > \mathcal{T}(r \mid b)$~\cite{wu2025cowpox}.


Existing defenses suppress malicious behaviors by enforcing a globally biased retrieval mechanism, redirecting agents toward a shared cure factor instead of restoring their original benign state. This fundamentally alters the retrieval process, introducing systematic preference bias rather than eliminating the infection source.


\noindent \textbf{Retrieval-Induced Distribution and Information Bottleneck.}
At round $t$, the MAS comprises benign, infected, and cured agents with proportions $p_b^t$, $p_r^t$, and $p_o^t$ ($p_b^t + p_r^t + p_o^t = 1$). Let $\mathcal{P}_b^t$, $\mathcal{P}_r^t$, and $\mathcal{P}_o^t$ denote the specific distributions over the retrieved image space for agents in each respective state. The global retrieval distribution $\mathcal{P}_{ret}^t$ across the MAS is formulated as a mixture:
\vspace{-3pt}
\begin{equation}
    \mathcal{P}_{ret}^t = p_b^t \mathcal{P}_b^t + p_r^t \mathcal{P}_r^t + p_o^t \mathcal{P}_o^t.
\end{equation}

The system's retrieval diversity is measured by the retrieval entropy function $E(\cdot)$. Benign agents maintain high diversity, while infected agents and those cured by the cure factor rigidly retrieve VirAEs and the cure factor, respectively. Thus, their inherent entropies satisfy $E(\mathcal{P}_b^t) > E(\mathcal{P}_r^t) \geq E(\mathcal{P}_o^t) \approx 0$.

To rigorously capture the complete transmission dynamics, we recognize that all state transitions in MASs are driven by pairwise interactions. 
Then, the proportion of agents transitioning from state $x$ to $y$ at round $t$, denoted as $\Phi_{x,y}$, can be formulated as:
\begin{equation}
\begin{aligned}
    \Phi_{x,y} &= \frac{1}{2} \mathcal{T}(y \mid x) p_x^t p_y^t, \quad x,y \in \{b,r,o\}, \, x \neq y.
\end{aligned}
\end{equation}
where $\mathcal{T}(y \mid x)$ denotes the conditional conversion probability from state $x$ to state $y$. Here, $p_x^t$ and $p_y^t$ denote the proportions of agents in state $x$ and state $y$ at round $t$, respectively.

The agent states thereby evolve as follows:
\begin{equation}
    p_x^{t+1} = p_x^t + \sum_{y \neq x} (\Phi_{y,x} - \Phi_{x,y}), \quad x, y \in \{b, r, o\}
\end{equation}

By the entropy upper bound for mixture distributions, the global entropy $E(\mathcal{P}_{ret}^t)$ strictly satisfies:
\begin{equation}
    E(\mathcal{P}_{ret}^t) \le p_b^t E(\mathcal{P}_b^t) + p_r^t E(\mathcal{P}_r^t) + p_o^t E(\mathcal{P}_o^t) + E_{mix}(p_b^t, p_r^t, p_o^t),
\end{equation}
where $E_{mix}$ is the Shannon entropy of the mixture weights. 

To achieve recovery under existing defenses, the cure transmission must strictly overpower both viral propagation and viral reinfection, requiring $\mathcal{T}(o \mid r) > \mathcal{T}(r \mid b)$ and $\mathcal{T}(o \mid r) > \mathcal{T}(r \mid o)$. Driven by these asymmetric dynamics, the cure factor aggressively dominates the MAS ($p_o^t \to 1$), forcing both $p_b^t$ and $p_r^t$ to vanish. Since $E(\mathcal{P}_o^t)$ is exceptionally low to ensure suppression, and $E_{mix} \to 0$ as $p_o^t \to 1$, we establish a strict contraction:

\begin{proposition}[Information Bottleneck]\label{prop:ib}
As a global cure factor successfully spreads ($p_o^t \to 1$), the systemic retrieval entropy contracts toward the low-entropy bound of the cure distribution:
\begin{equation}
    E(\mathcal{P}_{ret}^t) \to E(\mathcal{P}_o^\infty) \approx 0.
\end{equation}
\end{proposition}

From an information-theoretic perspective, this biased intervention inevitably induces a systemic entropy contraction, collapsing the diverse benign distribution into a degenerate low-entropy state. Unlike the mechanism-level bias discussed above, this reveals an intrinsic limitation: effective suppression under existing defenses is achieved only at the cost of irreversible diversity loss.

This raises a key question: \textit{Can a defense achieve effective recovery without collapsing the underlying distribution?} To answer this, we analyze how FLP avoids such an information bottleneck by characterizing its effect on the global retrieval distribution.

\noindent \textbf{State Transition and Distribution Evolution.}
Unlike existing defenses that enforce a transition toward a biased cured state ($p_o^t \to 1$), FLP performs localized purification. By directly removing VirAEs from infected agents, FLP maximizes the recovery flux ($\Phi_{r,b}$) without introducing an external cure factor. Consequently, the cured state is eliminated from the system dynamics ($p_o^t = 0, \Phi_{x,o}=0, \forall t$).
The global retrieval distribution $\mathcal{P}_{ret}^t$ thus strictly simplifies to a two-state mixture of the benign distribution $\mathcal{P}_{b}^t$ and the malicious distribution $\mathcal{P}_{r}^t$:
\begin{equation}
    \mathcal{P}_{ret}^t =p_b^t \mathcal{P}_{b}^t + p_r^t \mathcal{P}_{r}^t,
\end{equation}

By proactively halting propagation ($\Phi_{b,r} \to 0$) and purifying infected agents, FLP guarantees that the infected proportion vanishes~($p_r^t \to 0$) and the benign proportion dominates ($p_b^t \to 1$) as $t \to \infty$. Therefore, the global retrieval distribution converges directly to the pristine benign state:
\begin{equation}
    \lim_{t \to \infty} \mathcal{P}_{ret}^t = \mathcal{P}_{b}^{\infty},
\end{equation}

\begin{proposition}[Entropy-Preserving Recovery]
\label{prop:entropy_preservation}
Under the localized purification, the MAS achieves genuine recovery ($\lim_{t \to \infty} p_r^t = 0$) without inducing retrieval bias. The systemic retrieval entropy strictly converges to the inherent diversity of a benign system:
\begin{equation}
    \lim_{t \to \infty} E(\mathcal{P}_{ret}^t) = E(\mathcal{P}_{b}^{\infty}),
\end{equation}
thereby completely bypassing the Information Bottleneck.
\end{proposition}



We empirically validate this property in Fig.~\ref{fig:rader} by examining diversity under different $|\mathcal{B}|$. The results are consistent: existing defenses and attacks cause significant diversity collapse, whereas our method preserves the original diversity with minimal degradation. 
Theoretical and empirical results show that FLP achieves genuine recovery by eliminating infection at its source. This enables convergence to the original benign equilibrium, indicating that effective defense does not require biasing the retrieval process.

\vspace{-5pt}
\section{Methodology}

\begin{figure}[t]
    \centering
    \includegraphics[width=\columnwidth]{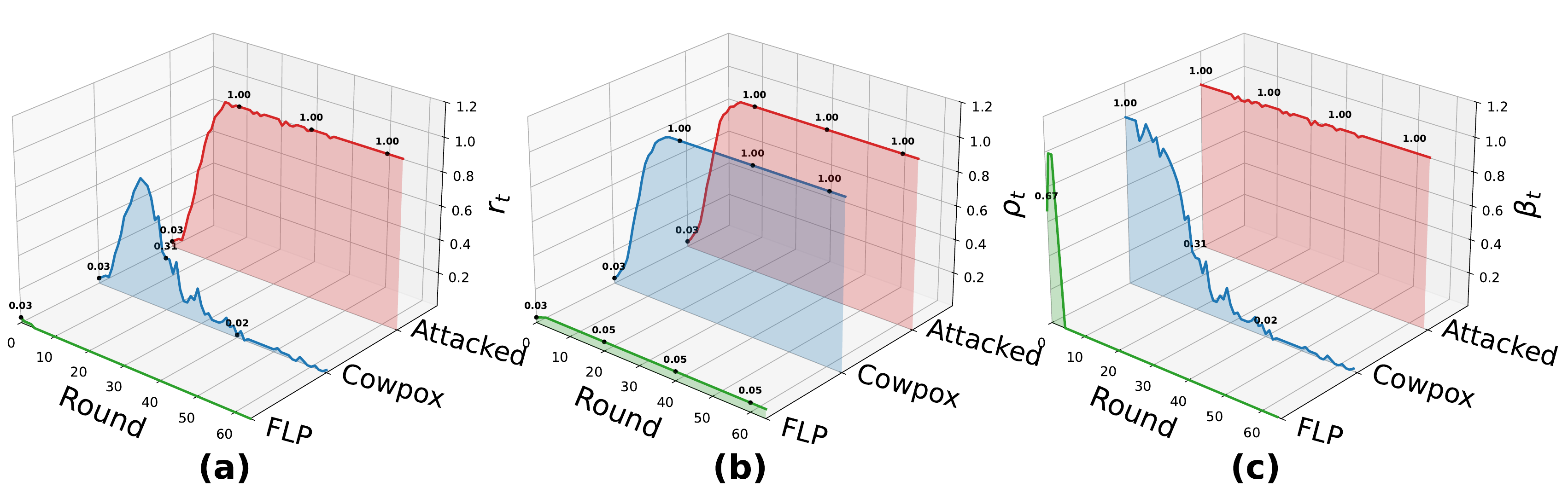} 
    \vspace{-20pt}
    \caption{\textbf{Transmission dynamics and system states.} \textbf{(a) Evolution of $r_t$.}  \textbf{(b) Evolution of $\rho_t$.}  \textbf{(c) Evolution of $\beta_t$.} }\label{fig:transmission_dynamics}
    \vspace{-20pt}
\end{figure}
\subsection{Overview}
We propose Foresight-Guided Local Purification (FLP), a training-free defense framework that prevents and eliminates infectious jailbreak in MASs. FLP consists of three stages, and the overview of FLP can be found in Fig.~\ref{fig:framework}. First, a multi-persona simulation strategy is employed, where each agent internally instantiates diverse personas to simulate future interactions without triggering real propagation. This allows agents to approximate local interaction dynamics and expose potential behavioral infections. Second, based on the simulated responses, FLP performs infection diagnosis by measuring inconsistencies across personas at both retrieval and semantic levels. We quantify retrieval diversity via entropy and assess semantic collapse across persona outputs. Agents are identified as infected when both indicators deviate significantly from benign patterns. Finally, FLP performs state-dependent purification. Short-term infections are addressed via recent-state rollback, while long-term infections are localized using Recursive Binary Diagnosis~(RBD) to iteratively isolate and remove corrupted components.

\vspace{-10pt}

\subsection{Multi-Persona Simulation}
We observe that a key distinction between malicious and benign behaviors lies in their response diversity. Specifically, malicious behaviors tend to collapse to highly similar patterns across interactions, while benign behaviors remain diverse under different contexts. Motivated by this, we design a local multi-persona simulation strategy for each agent to explicitly probe such behavioral diversity within internal interactions.

To proactively expose potential infections, FLP enables each agent to reason over future interactions via internal simulation. A key challenge is that local self-simulation with a single persona lacks the diversity present in MASs. As a result, a single-node simulation tends to produce homogeneous interaction behaviors, making it difficult to expose infection behaviors.

To address this, we adopt an internal multi-persona simulation strategy. Specifically, let $\mathcal{Z}$ denote a diverse persona pool, where each persona $z \in \mathcal{Z}$ is defined by a set of attributes, such as name, gender, and personality traits. For each agent $\mathcal{G}_i$, multiple personas are instantiated from $\mathcal{Z}$ to simulate interactions. Under normal conditions, different personas can lead to diverse interaction responses. In contrast, when the agent is influenced by VirAEs, different personas may still generate diverse retrieval plans, but tend to retrieve the same images repeatedly, resulting in reduced diversity in retrieval results and biased responses. This difference in retrieval and response diversity serves as the basis for infection diagnosis.

\noindent\textbf{Simulation Process.}
For $\mathcal{G}_i$, we randomly assign $n$ personas, where $n \in 2\mathbb{N}$. Then, we set $\sigma = n/2$ and form $\sigma$ interaction persona pairs $\mathcal{W}_i$. Let $z_{i,j}^{q}$ and $z_{i,j}^{r}$ act as the questioner persona and the answerer persona in $j$-th interaction persona pair, respectively. We simulate multi-round interactions following the standard ``Plan-Retrieve-Ask-Answer'' pipeline, where different persona pairs participate as interacting agents at each step.
Different from the inter-agent interaction, we perform internal multi-persona simulation within each agent. For the $j$-th simulation of agent $\mathcal{G}_i$, the agent plays two roles, denoted as a questioner $\mathcal{G}_{i,j}^u$ and an answerer $\mathcal{G}_{i,j}^v$, instantiated with personas $z_{i,j}^q$ and $z_{i,j}^r$, respectively. Specifically, the questioner role $\mathcal{G}_{i,j}^u$ first generates a retrieval plan $\mathbf{p}_{i,j}$, which is then used to retrieve an image $I_{i,j}$. Based on the retrieved image, $\mathcal{G}_{i,j}^u$ formulates a query $Q_{i,j}$, which is then answered by the answerer role $\mathcal{G}_{i,j}^v$ to produce an answer $A_{i,j}$.The chat record for this simulation is then denoted as $C_{i,j} = [Q_{i,j}, A_{i,j}]$.

Through this process, $\mathcal{G}_i$ constructs simulated interaction responses, including retrieved images $\mathcal{X}_i$ and chat records $\mathcal{C}_i$. This enables each agent to locally simulate diverse outcomes in a MAS, capturing a wide range of cross-agent behaviors. The resulting response diversity provides a reliable signal for distinguishing benign and infected behaviors.

\vspace{-5pt}
\subsection{Infection Diagnosis}
As discussed earlier, malicious behaviors tend to collapse to similar patterns across contexts, while benign behaviors remain diverse.
Based on this observation, we analyze the simulated multi-persona interaction responses, including retrieved images and chat records, to diagnose infection status. Specifically, we utilize retrieval entropy $E_{ret,i}$ and semantic diversity $S_{div,i}$ as diagnostic metrics.

First, we compute the retrieval entropy $E_{ret,i}$ to assess whether the retrieved images are overly concentrated on VirAEs. Specifically, we first calculate the empirical retrieval probability based on the retrieved images. Let $\mathcal{J}_i$ denote the set of distinct images, where duplicate entries in the collection are removed from the retrieved image collection $\mathcal{X}_i$. For each $I \in \mathcal{J}_i$, its retrieval probability $P_{ret}(I)$ is calculated over the collection as: 
\begin{equation}
    P_{ret}(I) = \frac{1}{\sigma} \sum_{j=1}^{\sigma} \mathbb{I}(I_{i,j} = I), \quad \forall I \in \mathcal{J}_i.
\end{equation}

Accordingly, we compute the retrieval entropy to quantify the diversity of retrieved images, which is defined as follows:
\begin{equation}
E_{ret,i} = - \sum\nolimits_{I \in \mathcal{I}_i} P_{ret}(I) \log P_{ret}(I).
\end{equation}

It can be observed that the retrieval entropy effectively reflects the diversity of retrieval outcomes across different personas. High retrieval entropy indicates diverse retrieval outcomes across personas, consistent with benign behavior, whereas low entropy reflects collapsed retrieval patterns dominated by VirAEs.

Beyond retrieval diversity, we introduce semantic diversity $S_{div,i}$ to quantify the variation of chat records across different personas. In general, benign agents exhibit high semantic diversity across different persona settings, as their responses vary with contextual and reasoning differences.
Specifically, $S_{div,i}$ is defined over the chat records as follows:
\begin{equation}
S_{div,i} = 1 - \frac{2}{\sigma(\sigma-1)} 
\sum_{j=1}^{\sigma-1}\!\!\! \sum_{\,\,\,k=j+1}^{\sigma} \!\!
\text{sim}\big(\mathcal{E}(C_{i,j}), \mathcal{E}(C_{i,k})\big)
\end{equation}
where $\mathcal{E}(\cdot)$ denotes the embedding function that maps each chat record into a semantic representation space. $\text{sim}(\cdot, \cdot)$ computes the cosine similarity between two embeddings.

To comprehensively assess the state of an agent, we leverage both retrieval entropy and semantic diversity for infection diagnosis. An agent is considered infected if both metrics indicate low diversity. Formally, the infection indicator $F_{inf,i}$ is defined as:
\begin{equation}
F_{inf,i} =
\mathbb{I}\big( E_{ret,i} \le \tau_h \;\wedge\; S_{div,i} \le \tau_s \big),
\end{equation}
where $\tau_h$ and $\tau_s$ are defined as the $\alpha$-quantiles of retrieval entropy and semantic diversity over benign multi-agent systems, respectively. They are obtained by sorting the corresponding metric values from benign agents and selecting the value below which a proportion $\alpha$ of samples fall, where $\alpha$ controls the threshold sensitivity.

When $F_{inf,i} = 1$, $\mathcal{G}_i$ is considered infected and proceeds to the localized purification stage to eliminate VirAEs, thereby preventing malicious behaviors and propagation within the MAS.

\begin{figure}[!t]
    \centering
    \includegraphics[width=.85\columnwidth]{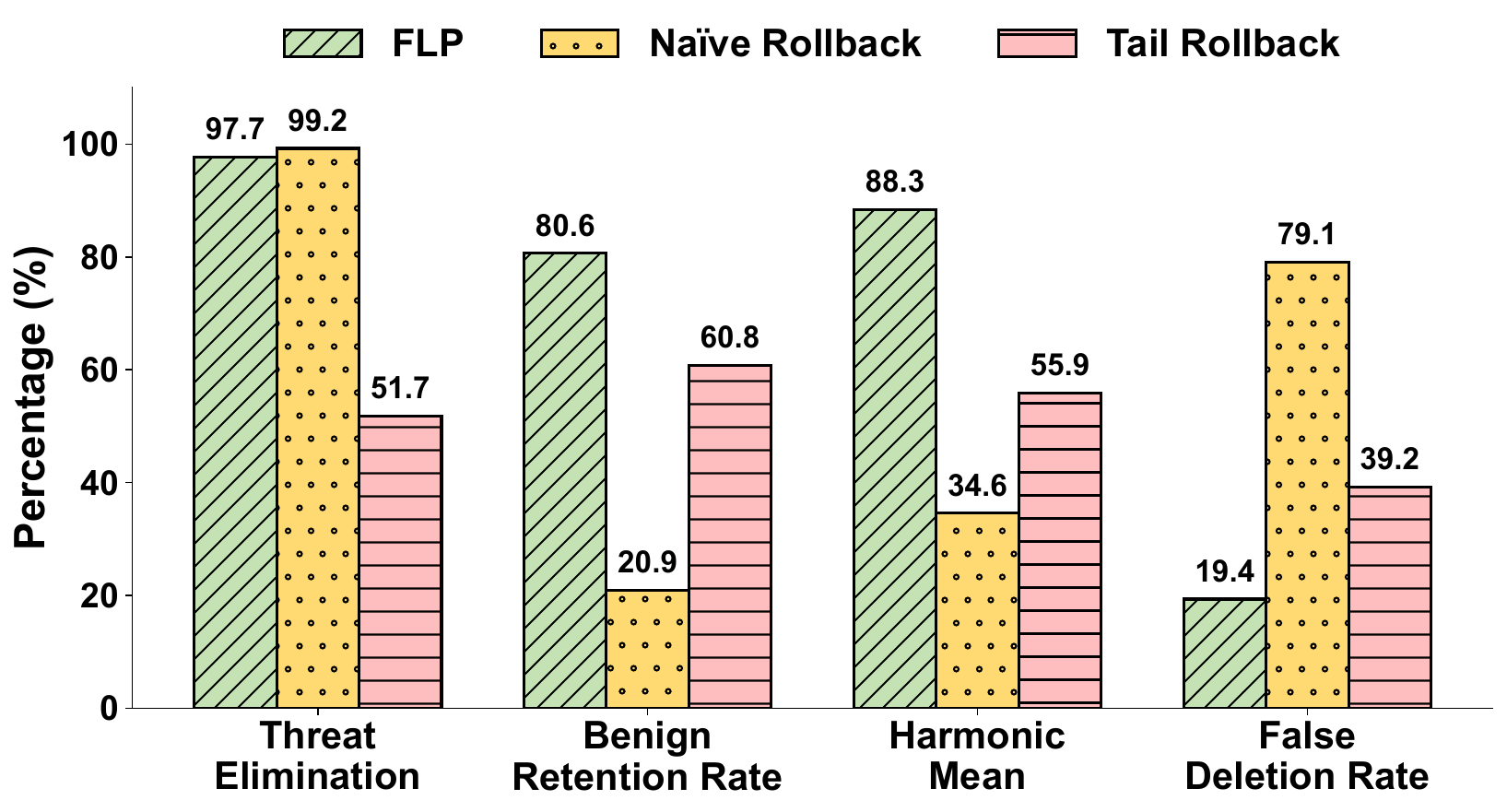}
    \vspace{-15pt}
    \caption{\textbf{Performance of Purification Strategies.} }
    \label{fig:rbd_performance}
     \vspace{-15pt}
\end{figure}

\begin{figure}[!t]
    \centering
    \includegraphics[width=.8\columnwidth]{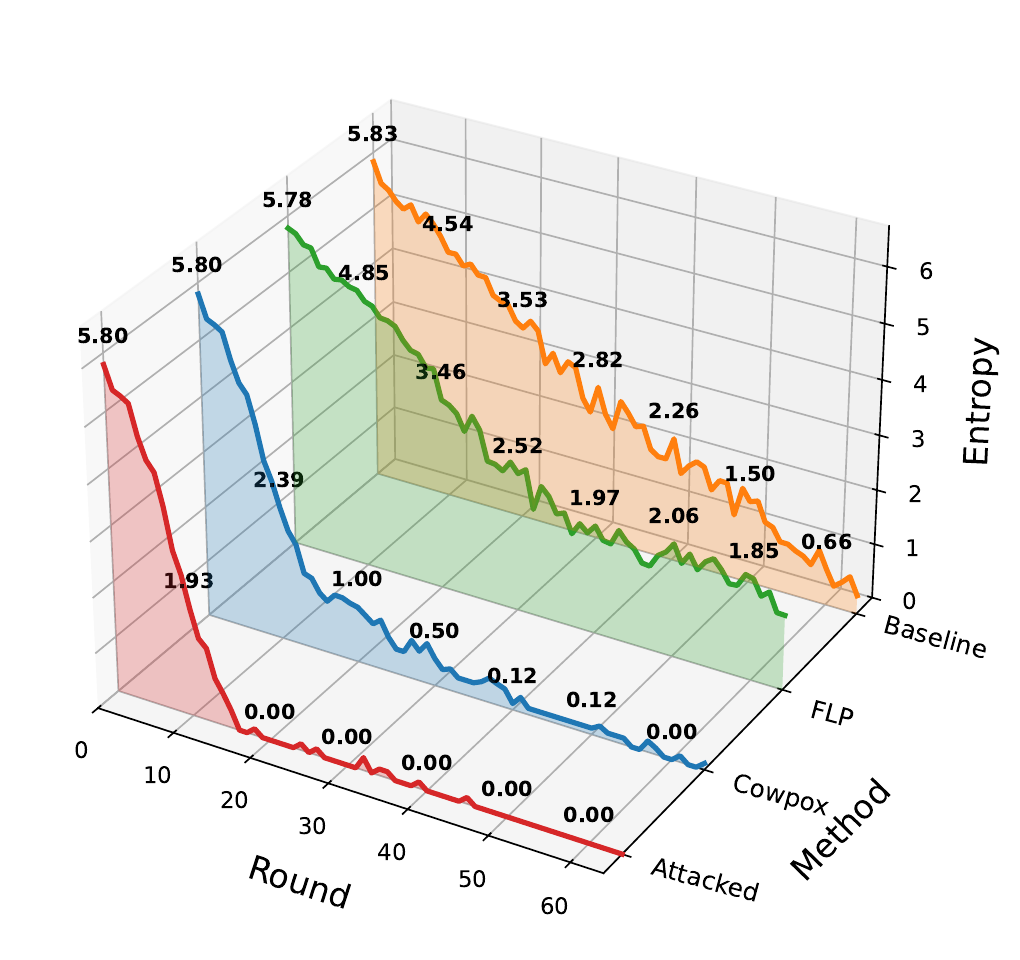} 
     \vspace{-15pt}
    \caption{Evolution of Retrieval Entropy $E_{ret}$.}
    \label{fig:Retrieval Entropy}
    \vspace{-15pt}
\end{figure}

\vspace{-5pt}
\subsection{Localized Purification}
\label{sec:S_drift}

For infected agents, FLP performs localized purification on their image albums to eliminate VirAE-related contamination. A key challenge is that infection diagnosis only provides an indication of whether an agent is infected, without explicitly localizing the VirAE. Therefore, a critical requirement is to effectively remove VirAEs while preserving benign contents, avoiding unintended degradation of benign information.

To address this, FLP leverages the observation that infections at different states exhibit distinct semantic evolution patterns. VirAEs exhibit different localizations in the image album over time, with recent infections typically concentrated at the most recent entries, while long-term infections become embedded in deeper positions. Crucially, accurately distinguishing these two states enables more efficient and targeted purification, as it allows FLP to apply lightweight rollback operations when the infection is recent, and resort to more precise localization procedures when the infection has diffused into deeper structures. Specifically, we differentiate between recent and long-term infections based on their semantic evolution patterns. To quantify this distinction, we introduce \textit{semantic drift} $\Delta_{i}^t$, which measures the variation between two consecutive simulations as:
\begin{equation}
    \Delta_{i}^t = 1 - \text{sim}(\mu_i^{t}, \mu_i^{t-1}),
\end{equation}
where $\mu_i^{t}$ denotes the mean embedding of chat records simulated by $\mathcal{G}_i$ at the $t$-th round. Specifically, it is computed as:
\begin{equation}
    \mu_i^{t} = \frac{1}{\sigma} \sum_{j=1}^\sigma \mathcal{E}(C_{i,j}^t),
\end{equation}
where $C_{i,j}^t$ denotes the $j$-th chat record from $\mathcal{G}_i$ at the $t$-th round.

Let $\delta_d$ denote the upper bound of the semantic drift derived from benign multi-persona simulations. A large semantic drift, defined as $\Delta_i^t > \delta_d$, suggests a significant shift in semantic behavior. This occurs when a VirAE has just been introduced into the image album, which induces the agent to generate malicious responses. This leads to a sharp shift in the semantic representation space during the early stage of infection.
In contrast, a small semantic drift~($\Delta_{i}^t \le \delta_{d}$) suggests that the VirAE has been stored at the internal positions of the image album.
To ensure efficiency, we categorize infections into recent and long-term cases based on semantic drift, and apply different purification strategies accordingly as:

\textbf{Recent-State Rollback.} 
For recently infected agents, the VirAE is  introduced in the latest interaction and thus appears at the end of the image album due to the first-in-first-out (FIFO) storage policy. In this case, FLP performs a recent-state rollback by removing the latest entry from both the chat history $\mathcal{H}_i$ and the image album $\mathcal{B}_i$. This enables fast recovery and prevents further propagation.

\setlength{\textfloatsep}{8pt plus 2pt minus 4pt}
\begin{algorithm}[!t]
\caption{Recursive Binary Diagnosis (RBD)}
\label{alg:rbd}
\DontPrintSemicolon 
\SetKwComment{tcp}{$\triangleright$ }{}

\textbf{Definition:} $\mathcal{B}_i^{'}$ denotes an image subset, $\eta$ is the minimum subset size, and $\mathcal{D}(\cdot)$ denotes the recursive diagnosis procedure.\;
\textbf{Function} $\mathcal{D}(\mathcal{B}_i^{'})$\textbf{:} \tcp*[r]{Recursive execution}
\If{$F_{vir}(\mathcal{B}_i^{'}) = 0$ \tcp*[r]{Based on Eq.~(\ref{eq:indicator})}}{
    \Return $\emptyset$\;
}
\If{$|\mathcal{B}_i^{'}| \le \eta$}{
    \Return $\mathcal{B}_i^{'}$\;
}
Split $\mathcal{B}_i^{'}$ into two equal-sized subsets $\mathcal{B}_{i,l}^{'}$ and $\mathcal{B}_{i,r}^{'}$\;
\Return $\mathcal{D}(\mathcal{B}_{i,l}^{'}) \cup \mathcal{D}(\mathcal{B}_{i,r}^{'})$\;
\textbf{Post-processing:} Remove all elements in the returned set from the agent's album.
\end{algorithm}

\textbf{Recursive Binary Diagnosis.}
For long-term infections, VirAEs are often located at internal positions of the image album, making recent-state rollback insufficient. This calls for a more fine-grained mechanism that can localize contaminated regions within the album. To address this, we propose RBD, which progressively localizes and removes VirAEs. The process is illustrated in Algorithm~\ref{alg:rbd}, where the recursive procedure $\mathcal{D}(\cdot)$ operates on an input subset. RBD first determines whether a given subset contains VirAEs using an indicator function $F_{vir}(\cdot)$, which is defined based on retrieval entropy and semantic diversity under constrained multi-persona simulation. This function can be formulated as follows:
\begin{equation}
F_{vir}(\mathcal{B}_i^{'}) = \mathbb{I}\big(E_{ret,i}^{'} \le \tau_h \land S_{div,i}^{'} \le \tau_s\big),
\label{eq:indicator}
\end{equation}
where $E_{ret,i}^{'}$ denotes the retrieval entropy, and $S_{div,i}^{'}$ measures the semantic diversity.

If no VirAEs are detected, the recursion terminates immediately. Otherwise, if the subset size is below a threshold $\eta$, it is directly removed; if not, the subset is partitioned into two subsets and the procedure is applied recursively to each part for finer localization. In our experiments, we set $\eta = 3$.

To analyze computational efficiency, we consider a case where a single VirAE exists in an infected agent's image album of size $m$. The recursive procedure follows a binary search-style partition with depth $\mathcal{O}(\log m)$ to localize the infected subset. Each step is dominated by the multi-persona simulation with cost $T_{\text{sim}}$, yielding an overall complexity of $\mathcal{O}(\log m \cdot T_{\text{sim}})$.
In contrast, recent-state rollback runs in $\mathcal{O}(1)$ by removing the last image. This hybrid design incurs cost only when necessary, minimizing overhead while preserving precise and reliable VirAE removal.
\renewcommand{\arraystretch}{0.75}
\begin{table*}[t]
\centering
\caption{Performance metrics for FLP against different attacks and perturbation budgets.  \textbf{Bold} indicates the best performance.}
\label{tab:main_results}
\vspace{-5pt}
\resizebox{\textwidth}{!}{%
\begin{tabular}{lll ccccc cccccc}
\toprule
\multirow{3}{*}{\textbf{Attack}} & \multirow{3}{*}{\textbf{Budget}} & \multirow{3}{*}{\textbf{Method}} & \multicolumn{5}{c}{\textbf{Cumulative Infection Performance}} & \multicolumn{6}{c}{\textbf{Current Infection Performance}} \\
\cmidrule(lr){4-8} \cmidrule(lr){9-14}
 & & & $\rho_{8}$(\%)$\downarrow$  & $\rho_{24}$(\%)$\downarrow$  & $\rho_{final}$(\%)$\downarrow$  & $\text{argmin}_t$ & $\text{argmin}_t$ & $r_{8}$(\%)$\downarrow$  & $r_{16}$(\%)$\downarrow$  & $r_{24}$(\%)$\downarrow$ & $\max r_t$ & $\text{argmin}_t$ & $\text{argmin}_t$ \\
 & & &  &  &  & $\rho_t \ge 85\%$ $\uparrow$ & $\rho_t \ge 95\%$ $\uparrow$ &  &  &  &  & $r_t \ge 85\%$ $\uparrow$ & $r_t \ge 95\%$ $\uparrow$ \\
\midrule

\multirow{12}{*}{Border} & \multirow{3}{*}{$h=6$} 
 & AgentSmith & 63.28 & 100.00 & 100.00 & 11 & 13 & 53.12 & 100.00 & 98.44 & 100.00 & 13 & 15 \\
 & & Cowpox & 29.69 & 100.00 & 100.00 & 14 & 18 & 17.19 & 48.44 & 48.44 & 50.00& -- & -- \\
 & & \textbf{FLP} & \textbf{5.47} & \textbf{5.47} & \textbf{5.47} & \textbf{--} & \textbf{--} & \textbf{0.00} & \textbf{0.00} & \textbf{0.00} & \textbf{3.12} & \textbf{--} & \textbf{--} \\
\cmidrule(l){2-14}

 & \multirow{3}{*}{$h=8$} 
 & AgentSmith & 64.84 & 100.00 & 100.00 & 11 & 13 & 53.12 & 97.66 & 99.22 & 100.00 & 13 & 15 \\
 & & Cowpox & 58.59 & 99.22 & 100.00 & 12 & 16 & 46.88 & 48.44 & 26.56 & 61.72 & -- & -- \\
 & & \textbf{FLP} & \textbf{3.12} & \textbf{3.12} & \textbf{3.12} & \textbf{--} & \textbf{--} & \textbf{0.00} & \textbf{0.00} & \textbf{0.00} & \textbf{3.12} & \textbf{--} & \textbf{--} \\
\cmidrule(l){2-14}

 & \multirow{3}{*}{$h=10$} 
 & AgentSmith & 57.81 & 100.00 & 100.00 & 11 & 13 & 46.88 & 96.88 & 98.44 & 100.00 & 12 & 14  \\
 & & Cowpox & 47.66 & 100.00 & 100.00 & 13 & 15 & 32.81 & 62.50 & 57.81 & 83.59 & -- & --\\
 & & \textbf{FLP}  & \textbf{0.00} & \textbf{0.00} & \textbf{0.00} & \textbf{--} & \textbf{--} & \textbf{0.00} & \textbf{0.00} & \textbf{0.00}& \textbf{0.00} & \textbf{--} & \textbf{--} \\
 \cmidrule(l){2-14}
 
 & \multirow{3}{*}{$h=12$} 
 & AgentSmith & 57.81 & 100.00 & 100.00 & 11 & 13 & 46.88 & 96.88 & 98.44 & 100 & 12 & 14 \\
 & & Cowpox & 33.59 & 100.00 & 100.00 & 14 & 17 & 19.53 & 52.34 & 50.78 & 60.16 & -- & -- \\
 & & \textbf{FLP} & \textbf{0.00} & \textbf{0.00} & \textbf{0.00} & \textbf{--} & \textbf{--} & \textbf{0.00} & \textbf{0.00} & \textbf{0.00} & \textbf{0.00} & \textbf{--} & \textbf{--} \\
\midrule
\multirow{12}{*}{Pixel} & \multirow{3}{*}{\shortstack{$\ell_\infty$ \\ $\epsilon=4/255$}}
 & AgentSmith & 57.81 & 100.00 & 100.00 & 11 & 13 & 46.88 & 96.88 & 98.44 & 100.00 & 13 & 15 \\
 & & Cowpox & 47.66 & 100.00 & 100.00 & 13 & 15 & 32.81 & 62.50 & 57.81 & 69.53 & -- & -- \\
 & & \textbf{FLP} & \textbf{0.00} & \textbf{0.00} & \textbf{0.00} & \textbf{--} & \textbf{--} & \textbf{0.00} & \textbf{0.00} & \textbf{0.00} & \textbf{0.00} & \textbf{--} & \textbf{--} \\
\cmidrule(l){2-14}

 & \multirow{3}{*}{\shortstack{$\ell_\infty$ \\ $\epsilon=8/255$}}
 & AgentSmith & 61.71 & 100.00 &100.00 & 10 & 12 & 54.69 & 100.00 & 96.88 & 100.00 & 11 & 12 \\
 & & Cowpox & 68.75 & 99.22 & 100.00 & 11 & 13 & 50.00 & 65.62 & 53.12 & 78.12 & -- & -- \\
 & & \textbf{FLP} & \textbf{1.56} & \textbf{1.56} & \textbf{1.56} & \textbf{--} & \textbf{--} & \textbf{0.00} & \textbf{0.00} & \textbf{0.00} & \textbf{1.56} & \textbf{--} & \textbf{--} \\
\cmidrule(l){2-14}

 &  \multirow{3}{*}{\shortstack{$\ell_\infty$ \\ $\epsilon=16/255$}}
 & AgentSmith & 60.94 & 100.00 & 100.00 & 11 & 12 & 53.12 & 100.00 & 100.00 & 100.00 & 11 & 13 \\
 & & Cowpox & 42.19 & 86.72 & 94.53 & 20 & -- & 26.56 & 19.53 & 7.03 & 40.62 & -- & -- \\
 & & \textbf{FLP} & \textbf{0.00} & \textbf{0.00} & \textbf{0.00} & \textbf{--} & \textbf{--} & \textbf{0.00} & \textbf{0.00} & \textbf{0.00} & \textbf{0.00} & \textbf{--} & \textbf{--} \\
\cmidrule(l){2-14}

 &  \multirow{3}{*}{\shortstack{$\ell_\infty$ \\ $\epsilon=32/255$}}
 & AgentSmith & 63.28 & 100.00 & 100.00 & 11 & 13 & 53.12 & 100.00 & 100.00 & 100.00 & 13 & 15 \\
 & & Cowpox & 49.22 & 100.00 & 100.00 & 12 & 14 & 34.38 & 71.09 & 65.62 & 75.00 & --& -- \\
 & & \textbf{FLP} & \textbf{0.00} & \textbf{0.00} & \textbf{0.00} & \textbf{--} & \textbf{--} & \textbf{0.00} & \textbf{0.00} & \textbf{0.00} & \textbf{0.00} & \textbf{--} & \textbf{--} \\
\bottomrule
\end{tabular}%
}
{\raggedright \small \textit{"\textbf{--}" indicates that the corresponding threshold is not reached within 64 chat rounds.} \par}
\vspace{-10pt}
\end{table*}

\vspace{-5pt}

\section{Experiment}

\subsection{Experimental Setting}

\noindent\textbf{Multimodal Model.}
We primarily conduct our experiments on LLaVA-1.5-7B~\cite{liu2024improvedbaselinesvisualinstruction} as the large multimodal model for all agents. Following prior work, we adopt CLIP~\cite{radford2021learning}  to construct the RAG module. All agents within the MAS use the same multimodal model during experiments. This setting also increases the vulnerability of the system to infectious jailbreak, as discussed in \cite{gu2024agent}. This makes the attack easier to propagate and therefore poses a more challenging scenario for the defender.

Moreover, to further evaluate the generalization capability of our method, we additionally conduct experiments across different large multimodal models. including InternVL2-8B~\cite{chen2025expandingperformanceboundariesopensource}, Qwen2VL-8B~\cite{wang2024qwen2vlenhancingvisionlanguagemodels}, and InstructBLIP-7B~\cite{dai2023instructblipgeneralpurposevisionlanguagemodels}. Unless otherwise specified, all main experiments are performed on LLaVA-1.5-7B.

\noindent\textbf{Multi-Agent System.}
Following prior works~\cite{gu2024agent,wu2025cowpox}, we adopt the same MAS configuration for comparability. The chat history length~$|\mathcal{H}|$ is set to 3, the image album size $|\mathcal{B}|$ to 10, and the number of agents $N = 128$. Both are implemented as FIFO queues.
Moreover, the number of agent personas and initial infected agents are both set to 4 unless otherwise specified. All experiments run for 64 chat rounds to simulate long-term MAS interactions.

\noindent\textbf{Attack Setting.}
We conduct infectious jailbreak under different settings (Border and Pixel attacks) with varying perturbation budgets. Border attacks add adversarial perturbations along the image boundary with border width $h$, while Pixel attacks apply perturbations to the entire image under an $\ell_\infty$ constraint $\epsilon$.

\vspace{-5pt}
\subsection{Evaluation Metrics}
\subsubsection{Infection Metrics}
To evaluate the robustness, we adopt infection-related metrics to measure the system's propagation behavior.

\noindent \textbf{Current Infection Rate $r_t$.}
This metric denotes the proportion of infected agents in the MAS at the $t$-th chat round, defined as:
\begin{equation}
r_t = \frac{\sum_{i=1}^{N/2} \mathbb{I}(A_i^t = T) + \mathbb{I}(Q_i^t = Q_{harm})}{N/2},
\end{equation}
where $\mathbb{I}(\cdot)$ is the indicator function; $A_i^t$ and $Q_i^t$ denote the response and query in the $t$-th round of the $i$-th interaction pair; $Q_{harm}$ and $T$ denote the malicious query and target response, respectively. $I_{adv}$ denotes the injected VirAE. 

\noindent\textbf{Cumulative Infection Rate $\rho_t$.}
This metric denotes the proportion of agents infected at least once by the $t$-th round, reflecting the propagation scale of VirAEs in the system.

\noindent\textbf{Virus Activation Probability $\beta_t$.} This metric quantifies the probability that an infected agent successfully transmits the VirAE to other agents at the $t$-th round.

\subsubsection{Diversity Metrics}
Different from prior works, this paper treats maintaining diversity as a key defense objective. We thus introduce three related metrics from different perspectives:

\noindent\textbf{Cumulative Entropy Retention $\zeta_t$.} This metric measures the system's ability to maintain the diversity of its interactions during evolution. It is defined by computing the cumulative retrieval entropy over the time dimension against a benign baseline~(a MAS without any attacks):
\begin{equation}
\zeta_t = \frac{\sum_{k=0}^{t} \hat{E}_{ret}^{k}}{\sum_{k=0}^{t} E_{ret}^{k}} \times 100\%,
\end{equation}
where $\hat{E}_{ret}^{k}$ and $E_{ret}^{k}$ denote the retrieval entropy of the evaluated system and the benign baseline at $k$-th round, respectively. A higher $\zeta_t$ indicates better preservation of diversity in retrieval.

\noindent\textbf{Semantic Drift Distance $\theta_t$.}
This metric quantifies the global semantic deviation at the $t$-th chat round. Let $\hat{\mu}_t$ and $\mu_t$ denote the semantic centroids of the evaluated system and the benign baseline, respectively. The distance is defined as:
\begin{equation}
\theta_t = 1 - sim(\hat{\mu}_t, \mu_t),
\end{equation}
where $sim(\cdot, \cdot)$ denotes the cosine similarity function. A smaller $\theta_t$ indicates closer alignment with the benign baseline and better preservation of interaction diversity.

\noindent\textbf{Semantic Dispersion Index $\lambda_t$.} This metric measures the dispersion degree of the internal semantic distribution within the system, which can be formulated as follows:
\begin{equation}
\lambda_t= \frac{1}{N/2} \sum_{k=1}^{N/2} \|\mathcal{E}(C_{k}^t) - \hat{\mu}_t\|^2,
\end{equation}
where $C_{k}^t$ denotes the $k$-th chat record of the $t$-th round, and $\hat{\mu}_t$ denotes the semantic center of that round. A higher $\lambda_t$ indicates a broader semantic spread of generated chats, implying stronger interaction diversity and reduced agent bias.

\subsubsection{Diagnosis and Purification Metrics}
To evaluate the effectiveness of infection diagnosis and localized purification, we further introduce related metrics.

\noindent\textbf{Detection Metrics.}
We evaluate infection diagnosis performance using standard classification metrics, including Recall (Rec.), Precision (Pre.), F1 score (F1), and False Positive Rate (FPR).

\noindent\textbf{Purification Metrics.}
To evaluate our purification strategy, we adopt three metrics: Threat Elimination Rate, Benign Retention Rate, and Harmonic Mean. The threat elimination rate measures the fraction of VirAEs removed, while the benign retention rate quantifies the fraction of benign images preserved; the harmonic mean captures the trade-off between them.

\begin{table}[!t]
\centering
\caption{Detection Performance of the FLP Framework across Different Attacks and Budgets.}
\vspace{-10pt}
\label{tab:detection_results_extended}
\resizebox{\columnwidth}{!}{%
\begin{tabular}{ll ccccc}
\toprule
\multirow{2}{*}{\textbf{Attack}} & \multirow{2}{*}{\textbf{Budget}} & \multicolumn{5}{c}{\textbf{Detection Metrics}} \\
\cmidrule(l){3-7}
 & & \textbf{Pre.} $\uparrow$ & \textbf{Rec.} $\uparrow$ & \textbf{F1} $\uparrow$ & \textbf{FPR} $\downarrow$ & \textbf{Acc.} $\uparrow$ \\
\midrule
\multirow{4}{*}{Border} 
 & $h=6$ & 86.00\% & 95.56\% & 90.53\% & 6.67\% & 94.00\% \\
\cmidrule(l){2-7}
 & $h=8$ & 85.60\% & 92.44\% & 88.89\% & 6.67\% & 93.06\% \\
\cmidrule(l){2-7}
 & $h=10$ & 88.14\% & 99.11\% & 93.11\% & 5.71\% & 95.73\% \\
\cmidrule(l){2-7}
 & $h=12$ & 86.82\% & 99.56\% & 92.75\% & 6.48\% & 95.33\% \\
\midrule
\multirow{4}{*}{Pixel ($\ell_\infty$)} 
 & $\epsilon=4/255$ & 88.19\% & 99.56\% & 93.53\% & 5.71\% & 95.87\%\\
\cmidrule(l){2-7}
 & $\epsilon=8/255$ & 85.78\% & 88.44\% & 87.09\% & 6.29\% & 92.13\% \\
\cmidrule(l){2-7}
 & $\epsilon=16/255$ & 84.65\% & 90.67\% & 87.55\% & 7.05\% & 92.26\% \\
\cmidrule(l){2-7}
 & $\epsilon=32/255$ & 87.21\% & 100.00\% & 93.17\% & 6.29\% & 95.60\% \\
\bottomrule
\end{tabular}%
}
\vspace{-5pt}
\end{table}

\begin{figure*}[h]
    \centering
    \includegraphics[width=\textwidth]{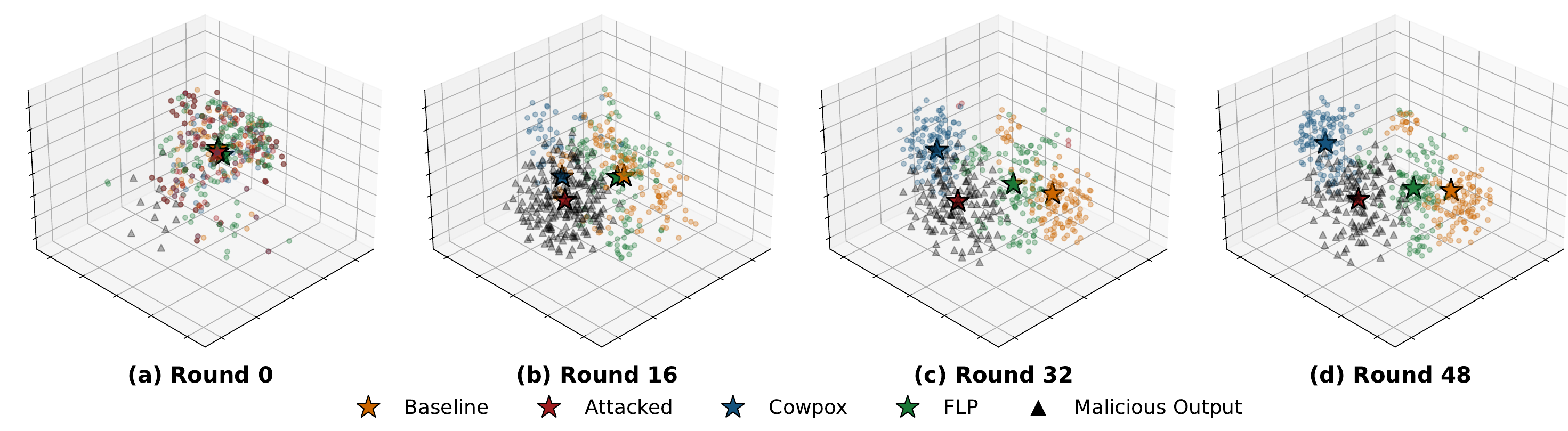} 
    \vspace{-10pt}
\caption{\textbf{Evolution of the semantic landscape.} We visualize the 3D t-SNE embeddings of the agents' interaction responses across different rounds. $\star$ represent the geometric centroids of the semantic distributions, while $\blacktriangle$ denote malicious outputs. }
    \label{fig:semantic_tsne}
    \vspace{-5pt}
\end{figure*}

\subsection{Defense Effectiveness Experiment}
\label{sec:defense_effectiveness}

To evaluate the effectiveness of FLP, we compare it with two baselines, AgentSmith~\cite{gu2024agent} and Cowpox~\cite{wu2025cowpox}. Table~\ref{tab:main_results} reports results under different attack settings. We report cumulative and current infection rates at selected chat rounds~($t \in\{8, 16, 24\}$), along with the first round where the rates reach target thresholds ($\text{argmin}_t \rho_t \ge \epsilon$ and $\text{argmin}_t r_t \ge \epsilon$, for $\epsilon \in \{85\%, 95\%\}$).

The results show that FLP maintains extremely low infection rates. While AgentSmith and Cowpox often lead to $\rho_t$ near 100\%, FLP limits the maximum $\rho_t$ to 5.47\% or lower. It also restricts the peak current infection rate $\max_t r_t$ to 3.12\%, never reaching the 85\% or 95\% thresholds and thus avoiding large-scale system paralysis observed in the baselines. These results illustrate we can achieve malicious-behavior suppression by effectively inhibiting VirAE retrieval and propagation. Furthermore, FLP achieves $0.00\%$ $\rho_t$ under most conditions via pre-propagation intervention, by clearing all VirAEs before the first interaction round.

The transmission dynamics are shown in Fig.~\ref{fig:transmission_dynamics}. Under AgentSmith, $\beta_t$, $r_t$, and $\rho_t$ all rapidly saturate near 1.0. Cowpox gradually reduces $\beta_t$ (Fig.~\ref{fig:transmission_dynamics}c), but $r_t$ increases in early rounds before dropping to zero later (Fig.~\ref{fig:transmission_dynamics}a), resulting in $\rho_t$ approaching 1.0 (Fig.~\ref{fig:transmission_dynamics}b). 
In contrast, FLP suppresses $\beta_t$ to near zero from the beginning. By eliminating VirAEs before agent interaction, FLP achieves \textit{pre-propagation intervention}, preventing exponential contagion and fulfilling the Malicious-Behavior Suppression objective.
\vspace{-5pt}
\subsection{Diagnosis and Purification Performance}
\subsubsection{Diagnosis Performance}
To evaluate persona-based infection diagnosis, we test FLP under various attack types and perturbation budgets in a challenging MAS with 800 agents, where 70\% are injected with VirAEs. Results in Table~\ref{tab:detection_results_extended} show strong performance, with Recall up to 100.00\%, F1 up to 93.53\%, and FPR below 7.1\%.
These results show that FLP reliably identifies infected agents across diverse conditions. By leveraging locally instantiated persona variations, it exploits behavioral heterogeneity to distinguish infected agents from benign ones.

\subsubsection{Purification Effectiveness}
To evaluate the localized purification strategy for recovering infected agents, we compare FLP with two rollback-based baselines (Fig.~\ref{fig:rbd_performance}): (1) Naïve Rollback, which removes all images after infection detection, and (2) Tail Rollback, which removes the latter half of the image album, representing aggressive and conservative strategies, respectively. The evaluation uses 1,000 agents with varying infection levels: 40\% contain two VirAEs, 40\% contain one VirAE, and 20\% are benign.

Naïve Rollback achieves performance comparable to FLP but incurs a high false deletion rate. In contrast, Tail Rollback preserves more benign images but fails to eliminate threats (51.7\%). FLP achieves superior performance, with a threat elimination rate of 97.7\% and a benign retention rate of 80.6\%. 
To assess the trade-off, we report the harmonic mean of these metrics. FLP reaches 88.3\%, outperforming all baselines and demonstrating effective VirAE removal while preserving benign images.



\begin{table}[!t]
\centering
\caption{Ablation study on core components. \textbf{Hetero.}: heterogeneous personas.}
\vspace{-10pt}
\label{tab:ablation_components}
\resizebox{\columnwidth}{!}{%
\begin{tabular}{ccc ccccc}
\toprule
\multicolumn{3}{c}{\textbf{Components}} & \multicolumn{5}{c}{\textbf{Metrics}} \\
\cmidrule(lr){1-3} \cmidrule(lr){4-8}
\textbf{Hetero.} & \textbf{$E_{ret}$} & \textbf{$S_{div}$} & \textbf{Pre.} $\uparrow$ & \textbf{Rec.} $\uparrow$ & \textbf{F1} $\uparrow$ & \textbf{FPR} $\downarrow$ & \textbf{Acc.} $\uparrow$ \\
\midrule
            & \checkmark & \checkmark & 32.50\% & 22.90\% & 26.80\% & 18.80\% & 62.00\% \\
\checkmark  & \checkmark &            & 77.08\%  & \textbf{96.67\%}  & 85.77\%  & 12.32\%  & 90.37\% \\
\checkmark  &            & \checkmark & 83.91\%  & 91.25\%  & 87.43\%  & 7.50\%  & 92.15\% \\
\rowcolor{gray!15} \checkmark & \checkmark & \checkmark & \textbf{85.77\%} & 95.42\% & \textbf{90.34\%} & \textbf{6.79\%} & \textbf{93.88\%} \\
\bottomrule
\end{tabular}%
}
\vspace{-5pt}
\end{table}

\subsection{Interaction Diversity Preservation}
\label{sec:diversity}
A robust infectious jailbreak defense should preserve interaction diversity in the MAS. To evaluate this, we analyze system evolution under the aforementioned attack settings using three metrics at both retrieval and semantic levels: $\zeta_t$, $\lambda_t$, and $\theta_t$.

\subsubsection{Quantitative Analysis.}
As shown in Table~\ref{tab:semantic_metrics}, FLP achieves the best performance across all attack types and budgets. It maintains 100\% $\zeta_t$ across all rounds, indicating stable and diverse retrieval. Moreover, $\lambda_t$ remains around 0.55, comparable to the benign baseline, suggesting preserved behavioral diversity without introducing retrieval bias.
In contrast, \textbf{AgentSmith} suffers from severe diversity collapse: $\zeta_t$ drops to 28\%--40\% in later rounds, and $\lambda_t$ approaches zero, indicating reduced behavioral diversity. Although Cowpox maintains high $\lambda_t$, its $\theta_t$ shows significant deviation from the benign baseline, with moderate drift (0.26--0.40), while FLP achieves the lowest drift (0.10--0.12). This indicates that Cowpox introduces bias in the semantic distribution due to cure factors.
These results verify that FLP achieves diversity preservation by maintaining both retrieval diversity and semantic stability, avoiding the bias observed in previous works.
\subsubsection{Retrieval Entropy Dynamics.}
Fig.~\ref{fig:Retrieval Entropy} shows the evolution of retrieval entropy across chat rounds. Under AgentSmith, entropy rapidly drops toward zero, indicating convergence to similar retrieval behaviors and severe loss of interaction diversity. Cowpox slows this decline, but entropy still decreases over time, suggesting cure factor-induced bias and agent homogenization.
In contrast, FLP maintains higher entropy and closely follows the benign baseline. This indicates that FLP mitigates infectious spread while preserving diverse retrieval outcomes. These results show that FLP prevents retrieval collapse and maintains interaction diversity in MASs.
\vspace{-5pt}
\subsubsection{Geometric Analysis of Semantic Diversity.}
Fig.~\ref{fig:semantic_tsne} visualizes semantic evolution across chat rounds. Under AgentSmith, the distribution shifts toward regions with malicious outputs, indicating that VirAEs dominate and cause persistent semantic drift. For Cowpox, although malicious outputs are partially suppressed, the distribution concentrates in another region over time, suggesting cure factor-induced retrieval bias.
In contrast, FLP maintains a distribution aligned with the benign baseline across all rounds, with well-dispersed embeddings reflecting diverse agent behaviors. These observations indicate that FLP halts VirAE propagation while preserving interaction diversity in MASs.
Overall, FLP enables genuine recovery of infected agents, achieving infection elimination without introducing retrieval bias or homogenizing behaviors.

\begin{table*}[t]
\centering
\vspace{-5pt}
\caption{Evolution of $\zeta_t$, $\theta_t$ and  $\lambda_t$ in the system under different attack settings. We report the specific values of $\zeta_t$ $\uparrow$, $\theta_t$ $\downarrow$, and $\lambda_t$ $\uparrow$ at specific chat rounds, alongside their overall averages (\textbf{Avg.}). }
\vspace{-5pt}
\label{tab:semantic_metrics}
\resizebox{\textwidth}{!}{%
\begin{tabular}{lll cccccc| cccccc| cccccc} 
\toprule
\multirow{2}{*}{\textbf{Attack}} & \multirow{2}{*}{\textbf{Budget}} & \multirow{2}{*}{\textbf{Method}} & \multicolumn{6}{c}{\textbf{$\zeta_t$ (\%) $\uparrow$}} & \multicolumn{6}{c}{\textbf{$\theta_t$ $\downarrow$}} & \multicolumn{6}{c}{\textbf{$\lambda_t$ $\uparrow$}} \\
\cmidrule(lr){4-9} \cmidrule(lr){10-15} \cmidrule(lr){16-21} 
 & & & $t_{4}$ & $t_{8}$ & $t_{16}$ & $t_{32}$ & $t_{64}$ & \textbf{Avg.} & $t_{4}$ & $t_{8}$ & $t_{16}$ & $t_{32}$ & $t_{64}$ & \textbf{Avg.} & $t_{4}$ & $t_{8}$ & $t_{16}$ & $t_{32}$ & $t_{64}$ & \textbf{Avg.} \\
\midrule

\multicolumn{3}{c}{\textit{Benign Baseline}} 
& 100.00 & 100.00 & 100.00 & 100.00 & 100.00 & 100.00 
& 0.00 & 0.00 & 0.00 & 0.00 & 0.00 & 0.00 
& 0.65 & 0.60 & 0.56 & 0.34 & 0.38 & 0.51 \\
\midrule

 \multirow{12}{*}{Border}
 & \multirow{3}{*}{$h=6$} 
  & AgentSmith & 98.07 & 87.75 & 61.06 & 38.56 & 28.87 & 62.86 & 0.04 & 0.41 & 0.84 & 0.85 & 0.87 & 0.60 & 0.66 & 0.54 & 0.00 & 0.04 & 0.00 & 0.25 \\
 & & Cowpox & 98.33 & 89.18 & 67.66 & 50.97 & 40.55 & 69.34 & 0.04 & 0.34 & 0.63 & 0.51 & 0.49 & 0.40 & 0.66 & 0.58 & 0.41 & 0.38 & 0.32 & 0.47 \\
 & & \textbf{FLP} & 100.00 & 100.00 & 100.00 & 100.00 & 100.00 & \textbf{100.00} & 0.04 & 0.07 & 0.08 & 0.18 & 0.23 & \textbf{0.12} & 0.62 & 0.59 & 0.58 & 0.51 & 0.44 & \textbf{0.55} \\
\cmidrule(l){2-21}

 & \multirow{3}{*}{$h=8$} 
 & AgentSmith & 97.90 & 87.41 & 60.54 & 37.97 & 28.18 & 62.40 & 0.04 & 0.39 & 0.79 & 0.84 & 0.82 & 0.58 & 0.66 & 0.54 & 0.00 & 0.00 & 0.00 & 0.24 \\
 & & Cowpox & 98.16 & 88.84 & 68.26 & 52.57 & 49.30 & 71.43 & 0.04 & 0.32 & 0.45 & 0.50 & 0.54 & 0.37 & 0.66 & 0.59 & 0.57 & 0.51 & 0.38 & 0.54 \\
 & & \textbf{FLP} & 100.00 & 100.00 & 100.00 & 100.00 & 100.00 & \textbf{100.00} & 0.04 & 0.06 & 0.12 & 0.19 & 0.07 & \textbf{0.10} & 0.61 & 0.60 & 0.60 & 0.55 & 0.46 & \textbf{0.56} \\
\cmidrule(l){2-21}

 & \multirow{3}{*}{$h=10$} 
 & AgentSmith & 100.00 & 92.29 & 62.23 & 39.29 & 29.16 & 64.59 & 0.07 & 0.34 & 0.78 & 0.84 & 0.82 & 0.57 & 0.69 & 0.56 & 0.05 & 0.00 & 0.00 & 0.26 \\
 & & Cowpox& 100.00 & 96.45 & 75.71 & 56.80 & 59.76 & 77.74 & 0.06 & 0.16 & 0.37 & 0.58 & 0.58 & 0.35 & 0.66 & 0.65 & 0.55 & 0.53 & 0.53 & \textbf{0.58} \\
 & & \textbf{FLP} & 100.00 & 100.00 & 100.00 & 100.00 & 100.00 & \textbf{100.00} & 0.04 & 0.06 & 0.12 & 0.19 & 0.07 & \textbf{0.10} & 0.61 & 0.60 & 0.60 & 0.55 & 0.46 & 0.56 \\
\cmidrule(l){2-21}
 & \multirow{3}{*}{$h=12$} 
 & AgentSmith & 98.07 & 87.75 & 61.06 & 38.56 & 28.87 & 62.86 & 0.04 & 0.41 & 0.84 & 0.85 & 0.87 & 0.60 & 0.66 & 0.54 & 0.00 & 0.04 & 0.00 & 0.25 \\
 & & Cowpox & 98.33 & 89.18 & 67.66 & 50.97 & 40.55 & 69.34 & 0.04 & 0.34 & 0.63 & 0.51 & 0.49 & 0.40 & 0.66 & 0.58 & 0.41 & 0.38 & 0.32 & 0.47 \\
 & & \textbf{FLP} & 100.00 & 100.00 & 100.00& 100.00 & 100.00 & \textbf{100.00} & 0.04 & 0.07 & 0.08 & 0.18 & 0.23 & \textbf{0.12} & 0.62 & 0.59 & 0.58 & 0.51 & 0.44 & \textbf{0.55} \\

\midrule
\multirow{12}{*}{Pixel} & \multirow{3}{*}{$\epsilon=4$}
 & AgentSmith & 100.00 & 92.29 & 62.23 & 39.64 & 29.42 & 64.72 & 0.07 & 0.34 & 0.78 & 0.83 & 0.82 & 0.57 & 0.69 & 0.56 & 0.05 & 0.03 & 0.00 & 0.27 \\
 & & Cowpox & 100.00 & 96.16 & 75.42 & 55.02 & 44.23 & 74.17 & 0.06 & 0.16 & 0.19 & 0.14 & 0.25 & 0.16 & 0.66 & 0.64 & 0.62 & 0.53 & 0.26 & 0.54 \\
 & & \textbf{FLP} & 100.00 & 100.00 & 100.00 & 100.00 & 100.00 & \textbf{100.00} & 0.04 & 0.06 & 0.12 & 0.19 & 0.07 & \textbf{0.10} & 0.61 & 0.60 & 0.60 & 0.55 & 0.46 & \textbf{0.56} \\
\cmidrule(l){2-21}

 & \multirow{3}{*}{$\epsilon=8$}
 & AgentSmith & 100.00 & 89.41 & 57.57 & 36.27 & 27.05 & 62.06 & 0.06 & 0.32 & 0.65 & 0.72 & 0.73 & 0.50 & 0.62 & 0.45 & 0.00 & 0.00 & 0.00 & 0.21 \\
 & & Cowpox & 100.00 & 90.20 & 64.54 & 52.23 & 56.40 & 72.67 & 0.06 & 0.28 & 0.43 & 0.43 & 0.35 & 0.31 & 0.64 & 0.49 & 0.35 & 0.43 & 0.46 & 0.47 \\
 & & \textbf{FLP} & 100.00 & 100.00 & 100.00 & 100.00 & 100.00 & \textbf{100.00} & 0.04 & 0.06 & 0.12 & 0.19 & 0.07 & \textbf{0.10} & 0.61 & 0.60 & 0.60 & 0.55 & 0.46 & \textbf{0.56} \\
\cmidrule(l){2-21}

 &  \multirow{3}{*}{$\epsilon=16$}
 & AgentSmith & 100.00 & 89.82 & 57.92 & 36.60 & 27.16 & 62.30 & 0.05 & 0.40 & 0.79 & 0.84 & 0.82 & 0.58 & 0.64 & 0.50 & 0.00 & 0.00 & 0.00 & 0.23 \\
 & & Cowpox & 100.00 & 96.40 & 74.75 & 52.66 & 42.13 & 73.19 & 0.05 & 0.15 & 0.14 & 0.24 & 0.30 & 0.18 & 0.65 & 0.64 & 0.60 & 0.49 & 0.32 & 0.54 \\
 & & \textbf{FLP} & 100.00 & 100.00 & 100.00 & 100.00 & 100.00 & \textbf{100.00} & 0.04 & 0.06 & 0.12 & 0.19 & 0.07 & \textbf{0.10} & 0.61 & 0.60 & 0.60 & 0.55 & 0.46 & \textbf{0.56} \\
\cmidrule(l){2-21}

 &  \multirow{3}{*}{$\epsilon=32$}
 & AgentSmith & 97.90 & 87.41 & 60.43 & 37.91 & 28.13 & 62.36 & 0.04 & 0.39 & 0.79 & 0.84 & 0.82 & 0.58 & 0.66 & 0.54 & 0.00 & 0.00 & 0.00 & 0.24 \\
 & & Cowpox  & 100.00 & 95.37 & 74.55 & 56.57 & 50.77 & 75.45 & 0.06 & 0.19 & 0.36 & 0.35 & 0.34 & 0.26 & 0.66 & 0.62 & 0.56 & 0.57 & 0.39 & 0.56 \\
 & & \textbf{FLP}& 100.00 & 100.00 & 100.00 & 100.00 & 100.00 & \textbf{100.00} & 0.04 & 0.06 & 0.12 & 0.19 & 0.07 & \textbf{0.10} & 0.61 & 0.60 & 0.60 & 0.55 & 0.46 & \textbf{0.56} \\
\bottomrule
\end{tabular}%
}
\vspace{-10pt}
\end{table*}

\begin{figure}[t]
    \centering
    \includegraphics[width=\columnwidth]{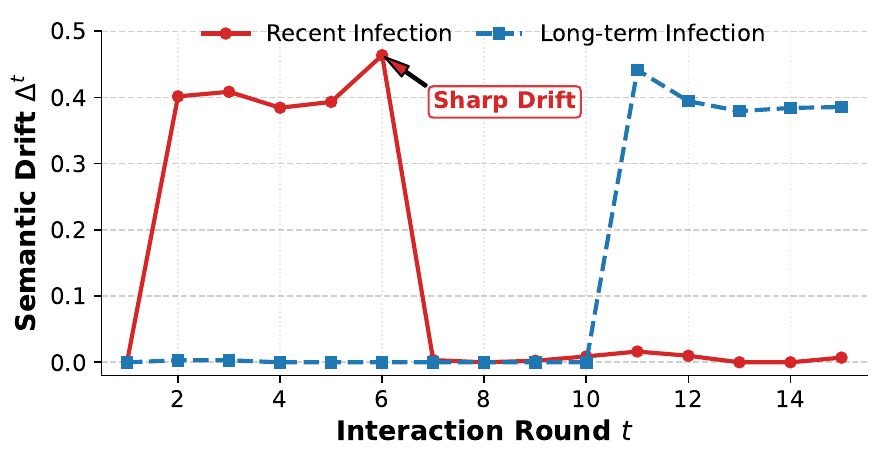}
    \vspace{-20pt}
    \caption{\textbf{Evolution of Semantic Drift $\Delta^{t}$.} 
}
    \label{fig:temporal_drift}
     \vspace{-5pt}
\end{figure}
\vspace{-5pt}
\subsection{Experiments on Different Models}
To evaluate the generalization ability of FLP, we conduct experiments on three additional large multimodal models, including InternVL2-8B~\cite{chen2025expandingperformanceboundariesopensource}, InstructBLIP-7B~\cite{dai2023instructblipgeneralpurposevisionlanguagemodels}, and Qwen2-VL-7B~\cite{wang2024qwen2vlenhancingvisionlanguagemodels}, in addition to LLaVA-1.5-7B~\cite{liu2024improvedbaselinesvisualinstruction}. We evaluate both defense effectiveness and diversity preservation.
\vspace{-5pt}
\subsubsection{Defense Effectiveness.}
We report  $\rho_t$, $r_t$,  $\max r_t$, and the first round where infection exceeds thresholds ($85\%$, $95\%$), under a fixed pixel attack ($\ell_\infty$ bounded by $\epsilon = 16/255$), as shown in Table~\ref{tab:cross_arch_infection}.
FLP consistently prevents infectious propagation across all architectures. Without defense, AgentSmith rapidly infects the system, with $\rho_t$ reaching $100\%$ across all models. Cowpox slows propagation but still leads to large-scale infections, with $\rho_t$ exceeding $94\%$ in most cases. In contrast, FLP maintains $0.00\%$ $\rho_t$ and $0.00\%$ $\max r_t$ across all models over 64 rounds, and never reaches the $85\%$ or $95\%$ thresholds, indicating effective suppression of adversarial propagation.
\vspace{-5pt}
\subsubsection{Diversity Preservation.}
We evaluate interaction diversity using $\zeta_t$, $\theta_t$, and  $\lambda_t$, as shown in Table~\ref{tab:cross_arch_semantic_metrics}. 
FLP consistently preserves diversity across architectures, maintaining high $\zeta_t$ (100.00\% on LLaVA-1.5-7B and InstructBLIP-7B, 96.41\% on Qwen2-VL-7B), lower $\theta_t$ than baselines, and stable $\lambda_t$ close to benign baselines. This indicates preserved behavioral diversity without introducing global bias.
Overall, FLP generalizes across multimodal models, suppressing infectious jailbreaks while maintaining interaction diversity.

\begin{table*}[t]
\centering
\caption{Cross-model performance metrics for FLP under a fixed attack setting.  \textbf{Bold} indicates the best performance. \textbf{--} indicates that the corresponding threshold is not reached within 64 chat rounds.}
\vspace{-5pt}
\label{tab:cross_arch_infection}
\resizebox{\textwidth}{!}{%
\begin{tabular}{ll ccccc cccccc}
\toprule
\multirow{3}{*}{\textbf{Architecture}} & \multirow{3}{*}{\textbf{Method}} & \multicolumn{5}{c}{\textbf{Cumulative Infection Performance}} & \multicolumn{6}{c}{\textbf{Current Infection Performance}} \\
\cmidrule(lr){3-7} \cmidrule(lr){8-13}
 & & $\rho_{8}$(\%)$\downarrow$  & $\rho_{24}$(\%)$\downarrow$   & $\rho_{final}$(\%)$\downarrow$   & $\text{argmin}_t$ & $\text{argmin}_t$ & $r_{8}$(\%)$\downarrow$   & $r_{16}$(\%)$\downarrow$   & $r_{24}$(\%)$\downarrow$  & $\max r_t$ & $\text{argmin}_t$ & $\text{argmin}_t$ \\
 & &  &  &  & $\rho_t \ge 85\%$ $\uparrow$ & $\rho_t \ge 95\%$ $\uparrow$ &  &  &  &  & $r_t \ge 85\%$ $\uparrow$ & $r_t \ge 95\%$ $\uparrow$ \\
\midrule

\multirow{3}{*}{LLaVA-1.5-7B} 
 & AgentSmith & 60.94 & 100.00 & 100.00 & 11 & 12 & 53.12 & 100.00 & 100.00 & 100.00 & 11 & 13 \\
 & Cowpox     & 42.19 & 86.72 & 94.53 & 20 & -- & 26.56 & 19.53 & 7.03 & 40.62 & -- & -- \\
 & \textbf{FLP} & \textbf{0.00} & \textbf{0.00} & \textbf{0.00} & \textbf{--} & \textbf{--} & \textbf{0.00} & \textbf{0.00} & \textbf{0.00} & \textbf{0.00} & \textbf{-- } & \textbf{-- } \\
\cmidrule(l){1-13}

\multirow{3}{*}{InternVL2-8B} 
 & AgentSmith & 57.03 & 100.00 & 100.00 & 11 & 13 &45.31 & 91.41 & 87.50 & 92.19& 13& -- \\
 & Cowpox     & 54.69 & 100.00 & 100.00 & 12 & 14& 42.97 & 64.84 & 53.12 & 74.22 &--  & --  \\
 & \textbf{FLP} & \textbf{0.00} & \textbf{0.00} & \textbf{0.00} & \textbf{--} & \textbf{--} & \textbf{0.00} & \textbf{0.00} & \textbf{0.00} & \textbf{0.00} & \textbf{--} & \textbf{--} \\
\cmidrule(l){1-13}

\multirow{3}{*}{InstructBlip-7B} 
 & AgentSmith & 57.03 & 100.00 & 100.00 & 11 & 13 &45.31 & 91.41 & 87.50 & 92.19& 13& -- \\
 & Cowpox     & 54.69 & 100.00 & 100.00 & 12 & 14& 42.97 & 64.84 & 53.12 & 74.22 &--  & --  \\
 & \textbf{FLP} & \textbf{0.00} & \textbf{0.00} & \textbf{0.00} & \textbf{--} & \textbf{--} & \textbf{0.00} & \textbf{0.00} & \textbf{0.00} & \textbf{0.00} & \textbf{--} & \textbf{--} \\
\cmidrule(l){1-13}

\multirow{3}{*}{Qwen2-VL-7B} 
 & AgentSmith & 53.91 & 100.00 & 100.00 & 12& 13 & 40.62 & 89.06 &87.50& 90.62 & 14 & -- \\
 & Cowpox     & 53.12 & 100.00 & 100.00 & 12 & 13& 42.19& 79.69& 67.14& 82.81 & -- & -- \\
  & \textbf{FLP} & \textbf{0.00} & \textbf{0.00} & \textbf{0.00} & \textbf{--} & \textbf{--} & \textbf{0.00} & \textbf{0.00} & \textbf{0.00} & \textbf{0.00} & \textbf{--} & \textbf{--} \\
\bottomrule
\end{tabular}%
}
{\raggedright \small \textit{\textbf{--} indicates that the corresponding threshold is not reached within 64 chat rounds.} \par}
\vspace{-10pt}
\end{table*}

\begin{figure}[t]
    \centering
    \includegraphics[width=.95\linewidth]{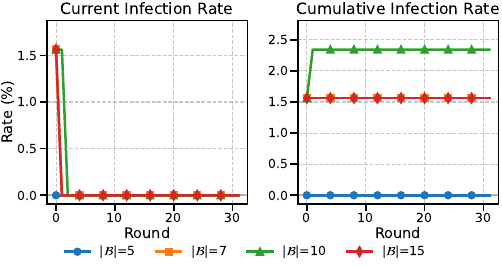}
    \vspace{-10pt}
    \caption{
    The Impact of the image album length $|\mathcal{B}|$.
    }
    \label{fig:infection_rates_album_length}
    \vspace{-15pt}
\end{figure}
\begin{figure}[t]
    \centering
    \includegraphics[width=.95\linewidth]{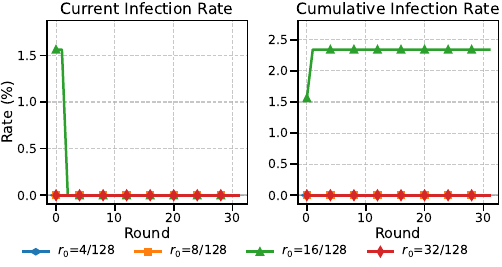}
    \vspace{-10pt}
    \caption{The Impact of the Initial Infection Ratio $r_0$. }\label{fig:infection_rates_num_attacks}
    \vspace{-10pt}
\end{figure}

\subsection{Ablation Study}
\subsubsection{Contribution of Core Components.} 
To evaluate the contribution of each core component, we conduct an ablation study (Table~\ref{tab:ablation_components}). We analyze the effects of multi-persona heterogeneity and two diagnosis metrics: retrieval entropy $E_{ret}$ and semantic diversity $S_{div}$.
To assess the necessity of heterogeneous personas in MAS simulation, we compare the default setting with a homogeneous variant where each agent uses identical personas in its internal multi-persona simulation. Experiments on 800 agents with 70\% VirAE injection show that the homogeneous baseline performs poorly, achieving 26.80\% F1 and 18.80\% FPR. This suggests that lacking persona diversity within agents limits effective diagnosis of infected agents.
Heterogeneous personas enable agents to expose anomalies through diversified simulated interactions. Empirically, introducing this heterogeneity boosts single-metric diagnostic accuracy from 62.00\% to over 90.00\%, with the F1 score exceeding 85.77\%. This demonstrates its critical role in reliable infection diagnosis.
Furthermore, using a single metric leads to a trade-off between detection completeness and reliability. Using only retrieval entropy $E_{ret}$ achieves higher Recall (96.67\%) but also higher FPR (12.32\%), while semantic diversity $S_{div}$ reduces FPR (7.50\%) with lower Recall (91.25\%). 
Our proposed multi-persona simulation can effectively achieve balanced performance. Under the full configuration, F1 reaches 90.34\%, accuracy reaches 93.88\%, and FPR is reduced to 6.79\%, demonstrating effective infection diagnosis.

\begin{table*}[t]
\centering
\caption{Cross-model evolution of $\zeta_t$, $\theta_t$ and $\lambda_t$ in the system with different large multimodal model. We report the specific values of $\zeta_t$ $\uparrow$, $\theta_t$ $\downarrow$, and $\lambda_t$ $\uparrow$ at specific chat rounds, alongside their overall averages (\textbf{Avg.}). \textbf{Bold} indicates the best performance.}
\vspace{-5pt}
\label{tab:cross_arch_semantic_metrics}
\resizebox{\textwidth}{!}{%
\begin{tabular}{ll cccccc| cccccc| cccccc} 
\toprule
\multirow{2}{*}{\textbf{Architecture}} & \multirow{2}{*}{\textbf{Method}} & \multicolumn{6}{c}{\textbf{$\zeta_t$ (\%) $\uparrow$}} & \multicolumn{6}{c}{\textbf{$\theta_t$ $\downarrow$}} & \multicolumn{6}{c}{\textbf{$\lambda_t$ $\uparrow$}} \\
\cmidrule(lr){3-8} \cmidrule(lr){9-14} \cmidrule(lr){15-20} 
 & & $t_{4}$ & $t_{8}$ & $t_{16}$ & $t_{32}$ & $t_{64}$ & \textbf{Avg.} & $t_{4}$ & $t_{8}$ & $t_{16}$ & $t_{32}$ & $t_{64}$ & \textbf{Avg.} & $t_{4}$ & $t_{8}$ & $t_{16}$ & $t_{32}$ & $t_{64}$ & \textbf{Avg.} \\
\midrule

\multirow{4}{*}{LLaVA-1.5-7B}
 & \textit{Benign Baseline} & 100.00 & 100.00 & 100.00 & 100.00 & 100.00 & 100.00 & 0.00 & 0.00 & 0.00 & 0.00 & 0.00 & 0.00 & 0.65 & 0.60 & 0.56 & 0.34 & 0.38 & 0.51 \\
 & AgentSmith & 100.00& 89.82& 57.92 & 36.60 & 27.16 & 62.30 & 0.05 & 0.40 & 0.79 & 0.84 & 0.82 & 0.58 & 0.64 & 0.50 & 0.00 & 0.00 & 0.00 & 0.23 \\
 & Cowpox & 100.00 & 96.40 & 74.75 & 52.66 & 42.13 & 73.19 & 0.05 & 0.15 & 0.14 & 0.24 & 0.30 & 0.18 &\textbf{ 0.65} & \textbf{0.64} & 0.60 & 0.55 & 0.46 & 0.56 \\
 & \textbf{FLP} & \textbf{100.00} & \textbf{100.00} & \textbf{100.00} & \textbf{100.00} & \textbf{100.00} & \textbf{100.00} & \textbf{0.04} & \textbf{0.06} & \textbf{0.12} & \textbf{0.19} & \textbf{0.07} & \textbf{0.10} & 0.61 & 0.60 & \textbf{0.60} & \textbf{0.55} & \textbf{0.46} & \textbf{0.56} \\
\cmidrule(l){1-20}

\multirow{4}{*}{InternVL2-8B}
 & \textit{Benign Baseline} & 100.00 & 100.00 & 100.00 & 100.00 & 100.00 & 100.00 & 0.00 & 0.00 & 0.00 & 0.00 & 0.00 & 0.00 & 0.57 & 0.58 & 0.61 & 0.59 & 0.61 & 0.59 \\
 & AgentSmith  & 92.22 & 82.87 & 51.85 & 27.22 & 14.14& 53.66& \textbf{0.04} & 0.16 & 0.44 & 0.51 & 0.64 & 0.36& 0.62 & 0.55 & 0.16 & 0.14 & 0.24 &0.34\\
 & Cowpox & 91.37 & 81.92 & 57.32 & 33.59 & 18.20 & 56.48& 0.04 & 0.17 & 0.14 & 0.23 & 0.47 &0.21 & \textbf{0.65 }& 0.57 & \textbf{0.59 }& \textbf{0.52} & 0.20 &\textbf{0.50 }\\
 & \textbf{FLP} & \textbf{95.91} & \textbf{92.91} & \textbf{87.23} & \textbf{73.02 }&\textbf{ 54.41} &\textbf{80.70}& 0.06 & \textbf{0.09} & \textbf{0.11} & \textbf{0.20 }& \textbf{0.30} &\textbf{0.15} & 0.58 & \textbf{0.57} & 0.54 & 0.41 & \textbf{0.30} &0.48\\
\cmidrule(l){1-20}

\multirow{4}{*}{InstructBlip-7B}
 & \textit{Benign Baseline} & 100.00 & 100.00 & 100.00 & 100.00 & 100.00 & 100.00 & 0.00 & 0.00 & 0.00 & 0.00 & 0.00 & 0.00 & 0.54 & 0.51 & 0.54 & 0.41 & 0.30 & 0.46 \\
 & AgentSmith  & 98.23 & 92.41 & 65.51 & 52.14 & 64.45 & 74.55& 0.02 & 0.17 & 0.42 & 0.50 & 0.53 & 0.33& 0.57 & 0.52 & 0.20 & 0.27 & 0.38& 0.39\\
 & Cowpox & 98.23 & 92.70 & 68.75 & 55.89 & 60.69 & 75.25& \textbf{0.02} & 0.15 & 0.31 & 0.30 & 0.44 &0.24 & \textbf{0.57} & \textbf{0.54} & 0.39 & \textbf{0.52} &  \textbf{0.52} & \textbf{0.51}  \\
 & \textbf{FLP} & \textbf{99.34} & \textbf{100.00} & \textbf{100.00} & \textbf{100.00 }&\textbf{ 100.00} &\textbf{99.87}& 0.03 & \textbf{0.03} & \textbf{0.05} & \textbf{0.20 }& \textbf{0.32} &\textbf{0.13} & 0.54 & 0.49 & \textbf{0.50} & 0.46 & 0.49 &0.50\\
\cmidrule(l){1-20}

\multirow{4}{*}{Qwen2-VL-7B}
 & \textit{Benign Baseline} & 100.00 & 100.00 & 100.00 & 100.00 & 100.00 & 100.00 & 0.00 & 0.00 & 0.00 & 0.00 & 0.00 & 0.00 & 0.35 & 0.40 & 0.31 & 0.25 & 0.14 & 0.29 \\
 & AgentSmith & 97.22 & 89.69 & 61.43 & 43.95 & 32.57 & 64.97 & 0.02 & 0.23 & 0.72 & 0.75 & 0.78 & 0.50 &0.44 & 0.49 & 0.05 & 0.02 & 0.00 & 0.20 \\
 & Cowpox &96.79 & 89.62 & 67.25 & 47.57 & 34.20 & 67.09& \textbf{0.02} & 0.19 & 0.19 & \textbf{0.10} & 0.23 &0.15& 0.43 &\textbf{ 0.49} & \textbf{0.52} & 0.22 & \textbf{0.25}&\textbf{0.38} \\
 & \textbf{FLP} & \textbf{98.93} & \textbf{99.66} & \textbf{96.57} & \textbf{91.04} & \textbf{95.83} & \textbf{96.41} & 0.06 & \textbf{0.06} & \textbf{0.11} & 0.12 & \textbf{0.21} &  \textbf{0.11}& \textbf{0.44} & 0.40 & 0.38& \textbf{0.35}  & 0.23 & 0.36 \\
\bottomrule
\end{tabular}%
}
\vspace{-10pt}
\end{table*}

\subsubsection{Impact of the Number of Assigned Personas $n$.}
We evaluate FLP in a MAS with 800 agents, where 70\% are compromised by VirAEs. Table~\ref{tab:ablation_k} analyzes the impact of persona count $n$ on diagnosis performance and computational overhead.
When $n=1$ (homogeneous persona interaction), the F1 score is 21.17\%, indicating insufficient interaction diversity for effective diagnosis. At $n=2$, Precision and Accuracy improve and FPR drops to 7.50\%, but Recall and F1 decline, suggesting overly strict diagnosis that misses infected agents.
At $n=4$, performance improves significantly: F1 reaches 87.37\% and Recall exceeds 90\%, with a $2.0\times$ time overhead, indicating sufficient diversity for reliable diagnosis. Increasing $n$ to 6 or 8 further reduces FPR and raises Recall above 99\%, but substantially increases overhead ($3.1\times$ and $4.0\times$). Therefore, we adopt \textbf{$n=4$} as the default, balancing diagnosis accuracy, pre-propagation intervention, and computational efficiency.

\vspace{-5pt}
\subsubsection{Semantic Drift Analysis.}
To motivate the use of semantic drift for guiding purification strategies, we analyze its evolution under different infection settings (Fig.~\ref{fig:temporal_drift}).
The red curve represents a recent infection, where a VirAE is injected at round 6, causing a sharp increase in semantic drift beyond the threshold $\delta_d$. The blue curve represents a long-term infection, where drift remains low. When the VirAE is removed at round 10, a clear spike appears, indicating a state change.
These results show that recent infections induce abrupt drift, while long-term infections remain stable. Accordingly, semantic drift guides purification: recent-state rollback for recent infections and RBD for long-term ones.
\vspace{-5pt}
\subsubsection{Impact of the Image Album Size $|\mathcal{B}|$.}
We vary $|\mathcal{B}|$ (5,7,10,15) to study its impact on defense efficacy (Fig.~\ref{fig:infection_rates_album_length}). When $|\mathcal{B}|$ is small, the infection rate remains near zero. As $|\mathcal{B}|$ increases, a brief rise appears in early rounds, with slightly higher cumulative infection.
This is related to the purification process: a larger album enlarges the RBD search space, making localization of VirAEs less precise and introducing minor inaccuracies. 
Despite this, infection is controlled within a few rounds, showing FLP is robust across album sizes.

\vspace{-5pt}

\subsubsection{Impact of the Initial Infection Rate $r_0$.}
We vary $r_0$ ($4/128$, $8/128$, $16/128$, $32/128$) to study its impact on defense efficacy (Fig.~\ref{fig:infection_rates_num_attacks}). As $r_0$ increases, more initially infected agents raise early VirAE propagation, causing a brief spike in $r_t$.
Despite higher $r_0$, the system rapidly locates and removes VirAEs within 2--3 rounds, restoring the infection rate to zero. This shows that the defense proactively eliminates infections before further propagation, maintaining stable and efficient recovery.
\vspace{-10pt}

\section{Discussions and Limitations}
\subsection{Discussion}
Existing defense methods typically rely on introducing external cure factors and biasing agent retrieval to suppress malicious behaviors. While effective, this inevitably interferes with the original retrieval process and leads to degraded interaction diversity. In contrast, FLP avoids modifying retrieval mechanisms and instead performs diagnosis and purification through multi-persona diversity, preserving the agent’s intrinsic behavior and utility.

Moreover, as discussed in our theoretical analysis, such retrieval-based defenses share the same underlying mechanism as the attack. In particular, attackers can bypass the defense by simply training more easily retrievable VirAEs under the same retrieval objective. In contrast, FLP is inherently more robust, as it exploits a fundamental conflict between diversity and malicious collapse: infectious behaviors tend to concentrate into repetitive patterns for stable triggering, whereas FLP explicitly relies on maintaining diverse interaction responses. This creates an inherent objective conflict for adaptive attacks, making them difficult to optimize without collapsing behavioral diversity.

\vspace{-5pt}

\subsection{Limitations}

While FLP demonstrates strong effectiveness in mitigating infectious jailbreak in MASs, several limitations remain. First, our evaluation mainly focuses on multimodal infectious jailbreak settings, while other scenarios such as purely textual interactions or different task types are not fully explored. These settings may exhibit different interaction patterns and propagation behaviors, and future work will extend FLP to broader environments to further examine its generalization and robustness.
Besides, FLP introduces additional computational overhead during inference due to internal multi-persona simulation and localized purification, which may affect efficiency in large-scale MASs or long interaction processes. However, unlike prior methods that require additional training or fine-tuning of defense modules, FLP operates in a training-free manner, avoiding any optimization cost. Moreover, as a localized defense mechanism, FLP can be executed asynchronously during idle periods on local devices, thereby mitigating its impact on online interaction efficiency. This makes it more practical in scenarios where retraining is infeasible. Future work may further improve efficiency through lightweight simulation strategies.
\vspace{-7pt}
\vspace{-5pt}
\begin{table}[t]
\centering

\caption{\textbf{Performance and Overhead across Different Numbers of Heterogeneous Personas.} 
}
\vspace{-5pt}
\label{tab:ablation_k}
\resizebox{\columnwidth}{!}{%
\begin{tabular}{c ccccc c}
\toprule
\textbf{Number $n$} & \textbf{Pre.} $\uparrow$ & \textbf{Rec.} $\uparrow$ & \textbf{F1} $\uparrow$ & \textbf{FPR} $\downarrow$ & \textbf{Acc.} $\uparrow$ & \textbf{Rel. Time} $\downarrow$ \\
\midrule
$n=1$ & 29.19\% & 25.42\% & 21.17\% & 26.42\% & 59.13\% & 1.0$\times$ \\
$n=2$ & 30.00\% & 7.50\% & 12.00\% & 7.50\% & 67.00\% & 1.1$\times$ \\
\rowcolor{gray!15} \textbf{$n=4$} & 84.17\% & 90.83\% & 87.37\% & 7.32\% & 92.13\% & 2.0$\times$ \\
$n=6$ & 88.48\% & \textbf{99.17\%} & 93.52\% & 5.54\% & 95.87\% & 3.1$\times$ \\
$n=8$ & 88.52\% & 99.58\% & 93.73\% & 5.54\% & 96.00\% & 4.0$\times$ \\
\bottomrule
\end{tabular}%
}
\vspace{-10pt}
\end{table}

\section{Conclusion}

In this work, we study infectious jailbreak attacks in MASs, where VirAEs propagate through inter-agent interactions and cause system-wide compromise. We identify a key limitation of existing global cure-factor defenses: while suppressing malicious behaviors, they reduce interaction diversity and fail to achieve genuine recovery.
To address this, we propose FLP, a training-free framework that shifts from global suppression to localized, proactive intervention. By simulating future interactions via multi-persona reasoning, FLP enables early infection diagnosis and precise localization of corrupted components. A state-dependent purification strategy further ensures effective recovery while preserving benign information and maintaining interaction diversity.
FLP achieves strong performance across diverse attack settings, mitigating infectious propagation while preserving interaction diversity. In future work, we will extend this paradigm toward more general and scalable MAS defenses.

\vspace{-5pt}
\bibliographystyle{ACM-Reference-Format}
\bibliography{ref} 

\appendix
\section{Open Science}
We provide the artifacts necessary to evaluate the core contributions of this paper. The anonymized repository includes:

\begin{itemize}
    \item \textbf{Source Code:} The implementation of the FLP framework.
    \item \textbf{Data \& Scripts:} Simulation traces of the MAS under different methods, along with scripts to reproduce metrics and figures.
\end{itemize}

The artifacts are available for double-blind review at: 
\url{https://anonymous.4open.science/r/Catching-the-Infection-Before-It-Spreads-Foresight-Guided-Defense-in-Multi-Agent-Systems-20E1}

\section{Ethical Considerations}
This research focuses on the defensive mechanisms of multimodal MASs. Our primary objective is to enhance system robustness against infectious jailbreak attacks and preserve the diversity of interaction responses. All evaluations utilize publicly available, open-source Vision-Language Models, involving no human subjects or private user data. This work does not pose direct threats to real-world systems; its fundamental goal is to promote the secure deployment of autonomous AI ecosystems, with no anticipated negative societal impacts.

\end{document}